\documentclass[runningheads]{llncs}

\usepackage{eccv}

\usepackage{eccvabbrv}

\usepackage{graphicx}
\usepackage{tocloft}
\usepackage{booktabs}

\usepackage[accsupp]{axessibility}  

\usepackage{hyperref}

\usepackage{orcidlink}

\usepackage{algorithm}
\usepackage{algorithmic}
\usepackage{float}
\usepackage[table]{xcolor}
\def\ours{\texttt{Search2Motion}}
\def\search{\texttt{ACE-Seed}}

\def\davis{S2M-DAVIS}
\def\objmove{S2M-OMB}
\def\metrics{FLF2V-obj}

\definecolor{best}{RGB}{255,200,100}
\definecolor{second}{RGB}{255,243,194}
\begin{document}

\title{Search2Motion: Training-Free Object-Level Motion Control via Attention-Consensus Search} 

\titlerunning{Search2Motion}

\author{Sainan Liu*\inst{1}\and
Tz-Ying Wu*\inst{1} \and
Hector A Valdez*\inst{1} \and
Subarna Tripathi\inst{1}}

\institute{Intel Corporation}

\maketitle
\begingroup
\renewcommand\thefootnote{\textasteriskcentered}
\footnotetext{Equal contribution}
\endgroup

\begin{figure}
    \centering
    \includegraphics[width=\linewidth]{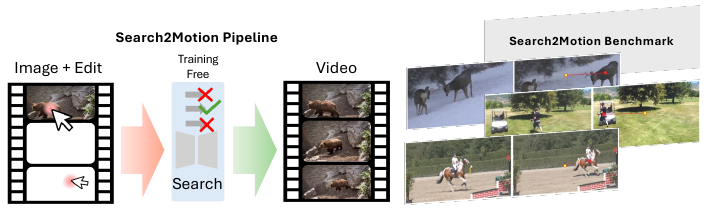}
    \vspace{-20pt}
    \caption{\textit{Left:} \ours{} is a training-free pipeline for object-level motion editing. Given a single image and a user-specified target location, \ours{} constructs a target frame and leverages pretrained FLF2V motion priors to synthesize realistic object motion, without retraining or auxiliary control signals. \textit{Right:} Sample pairs from the \ours{} Benchmark, two stable-camera datasets for object-only motion evaluation.}
    \label{fig:teaser}
    \vspace{-15pt}
\end{figure}
\begin{abstract}
  We present \ours{}, a training-free framework for object-level motion editing in image-to-video generation. Unlike prior methods requiring trajectories, bounding boxes, masks, or motion fields, \ours{} adopts target-frame-based control, leveraging first-last-frame motion priors to realize object relocation while preserving scene stability without fine-tuning. Reliable target-frame construction is achieved through semantic-guided object insertion and robust background inpainting. We further show that early-step self-attention maps predict object and camera dynamics, offering interpretable user feedback and motivating \search{} (Attention Consensus for Early-step Seed selection), a lightweight search strategy that improves motion fidelity without look-ahead sampling or external evaluators. Noting that existing benchmarks conflate object and camera motion, we introduce \davis{} and \objmove{} for stable-camera, object-only evaluation, alongside
  \metrics{} metrics that isolate object artifacts without requiring ground-truth trajectories. \ours{} consistently outperforms baselines on \metrics{} and VBench. Additional details are available in the \href{https://intelailabpage.github.io/2026/03/13/search2motion.html}{Project Page}.

  
    
\end{abstract}
\section{Introduction}
\label{sec:intro}
\vspace{-5pt}

Video generation models have advanced rapidly in recent years, delivering remarkably realistic and temporally coherent results. However, enabling users to reliably edit object motion in image-to-video generation remains a challenge. State-of-the-art open-source models are typically optimized for generic generation settings: text-to-video (T2V), image-to-video (I2V), or first--last-frame (FLF2V), which rarely support object-level motion editing capabilities as a first-class feature. When stronger controllability is needed, existing approaches introduce additional supervision, model-specific tuning, or auxiliary control modules that are tightly coupled to a particular backbone. Beyond portability concerns, these additions often degrade the model's native generation prior, compromising visual quality or temporal coherence. As new video generators are released rapidly, control mechanisms designed for one model quickly become obsolete, necessitating substantial re-engineering to adapt.


A further practical barrier is that end users rarely have access to the precise control signals assumed by these methods, such as per-frame depth maps, dense motion fields, bounding-box trajectories, or carefully curated keyframes. The most common starting point is simply a single input image and an intuitive intent: \emph{where should the object move?} Methods that depend on auxiliary signals are difficult to apply in real workflows, while approaches that synthesize intermediate constraints (\eg, edited intermediate frames) are often brittle and require repeated trial-and-error.


We introduce \ours{}, a training-free, modular pipeline for cut-and-paste object-motion editing in image-to-video generation, organized into three stages: \emph{target frame construction}, \emph{motion synthesis}, and \emph{seed selection}. Each stage is designed to be independently applicable, making the overall system both practical and model-agnostic. Our contributions are:

\begin{itemize}
    \item {\bf User-friendly motion control.} We reformulate object motion editing as a FLF2V generation task, allowing users to specify only a target object location suggested by the semantic-guided object placement pipeline, rather than designing a full trajectory.

    \item \textbf{Attention-based dynamics prediction.} We show that early-step diffusion self-attention maps predict object and camera dynamics, providing interpretable user feedback before committing to full generation.

    \item \textbf{ACE-Seed.} An attention-consensus noise-space search strategy that improves motion fidelity without look-ahead sampling, external evaluators, or heavy compute. \search{} also applies as a drop-in module in standard FLF2V settings beyond \ours{}.

    \item \textbf{Benchmark and metrics.} We introduce \davis{} and \objmove{} for stable-camera object-only evaluation, and \metrics{}, metrics that isolate object motion artifacts without ground-truth trajectories.
\end{itemize}

\begin{figure}[t]
    \centering
    \includegraphics[width=\linewidth]{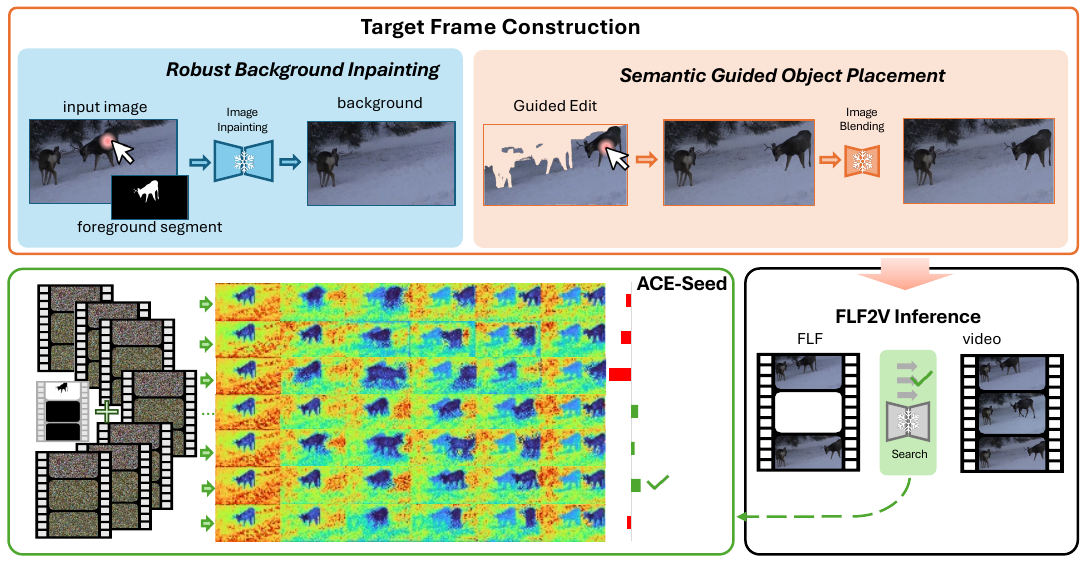}
    \vspace{-15pt}
    \caption{The \ours{} Pipeline is constructed with three components, where the user can interact with the application at the target frame construction stage (Background Inpainting and Object Placement). Then the original input image and the user-edited last frame are sent to a first-frame last-frame (FLF2V) video generator to generate the final video based on the given input image and user preference, while utilizing \search{}, a novel search criterion in the noise space based on attention-consensus, to improve the generated video quality automatically.}
    \label{fig:pipeline}
\end{figure}

\vspace{-20pt}


\section{Related Work}
\vspace{-5pt}

\subsection{Motion Control for Video Diffusion}
Existing controllable video generation methods largely fall into two families.

\subsubsection{Fine-Tuned Motion Guidance} approaches condition generation on explicit spatiotemporal controls, including trajectories~\cite{yin2023dragnuwa, wu2024draganything, wang2024motionctrl, zhang2025tora, zhou2025trackgo, singer2025ttm}, bounding boxes or masks~\cite{yariv2025throughthemask}, motion fields~\cite{shi2024motioni2v}, or fused multi-signal controllers~\cite{jiang2025vace}. These methods can deliver precise control, but typically require paired training data, additional training stages, or specialized control modules, which increases engineering and compute cost and can be difficult to port across rapidly evolving backbones. Even when training is not required \cite{singer2025ttm}, these interfaces often assume expert-designed control inputs (\eg carefully tuned trajectories), which are nontrivial for typical users to specify.
\vspace{-10pt}

\subsubsection{Frame-Based Guidance} is widely used in both proprietary and open-source systems, which steer motion by conditioning on a reference frame sequence (\eg sparse keyframes) ~\cite{runway_video_generation, kling_video, pikalabs, HaCohen2024LTXVideo, jiang2025vace}. When only a single first frame is available, practitioners often synthesize a keyframe using image-editing workflows or image diffusion models~\cite{machalek-2020-kontext, wu2025qwenimagetechnicalreport}. However, these workflows typically provide little assistance on \emph{where} an object can plausibly be placed. Users must manually explore placements through trial-and-error, whereas our semantic-guided insertion can automatically propose feasible candidate regions for users to select from. 
\vspace{-10pt}


\subsection{Inference-Time Search and Seed Selection}
Selecting diffusion noise seeds is challenging in video generation because intermediate latents are not human-interpretable and are difficult to score reliably. Recent inference-time optimization methods address this by performing one- or multi-step look-ahead: partially denoising candidates to produce draft frames/videos and scoring them using learned embeddings (\eg, VLM or DINO-style rewards)~\cite{ma2025inferencetimescalingdiffusionmodels, oshima2025inference, yang2025scalingnoisescalinginferencetimesearch}. Although effective, these strategies introduce sequential overhead and depend on external evaluators, limiting their suitability for interactive workflows. Concurrent work BANSA~\cite{kim2025modelknowsbestnoise} scores seeds via an entropy-based Bayesian metric from attention maps and validates on text-to-video models. In contrast, \search{} targets frame-conditioned generation (FLF2V) and selects seeds via early-step attention consensus to improve object-level motion fidelity.
\vspace{-10pt}

\subsection{Object-Level Evaluation: Metrics and Benchmarks}
\begin{figure}[t!]
    \centering    
    \begin{minipage}[h]{.51\linewidth}
        \includegraphics[width=\linewidth]{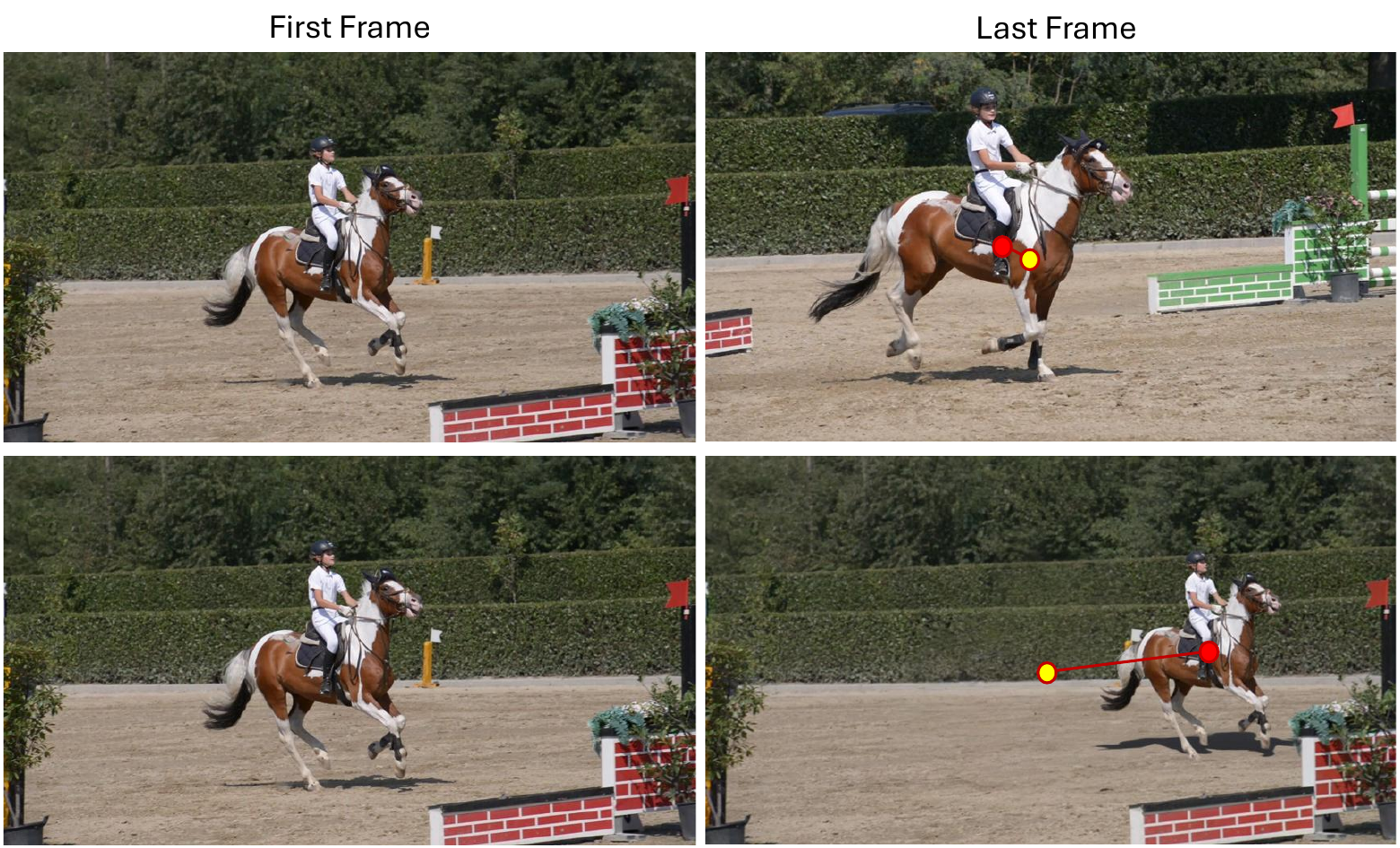}
    \caption{Object trajectory between first (yellow point) and last frame (red point). The upper pair of frames is extracted from raw video in \cite{DAVIS_Perazzi2016}, and the lower pair is from our synthesized dataset, \davis{}.
   }
    \label{fig:traj}
    \end{minipage}
    \hfill
    \begin{minipage}[h]{.47\linewidth}
        \vspace{15pt}
        \includegraphics[width=\linewidth]{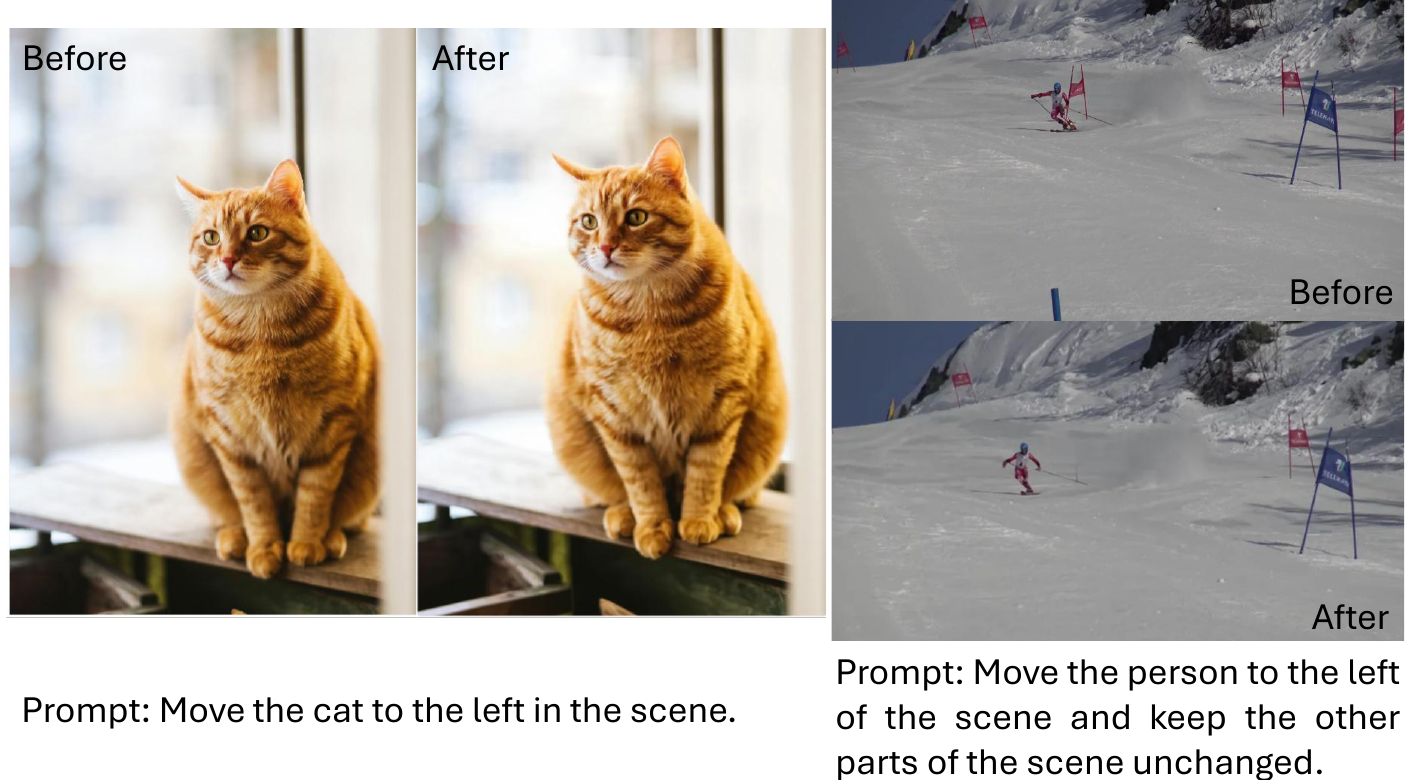}
        \captionof{figure}{Qualitative examples for object replacement using state-of-the-art image editing tools, Qwen-Image-Edit\cite{wu2025qwenimagetechnicalreport} (left) and FLUX-Kontext\cite{machalek-2020-kontext} (right).}
    \label{fig:placement}
    \end{minipage}
\end{figure}

\vspace{-10pt}
Object-motion assessment remains underdeveloped. Popular benchmarks such as VBench~\cite{huang2024vbench} primarily compute whole-frame metrics (\eg, CLIP-based alignment, aesthetics, temporal consistency, motion smoothness), which entangle foreground motion with background changes, and can miss artifact-prone object dynamics (jitter, drift, implausible trajectories). Prior motion-editing works~\cite{yin2023dragnuwa, wu2024draganything} often rely on qualitative comparisons or trajectory-based evaluation on videos where camera and object motion co-occur (\cref{fig:traj}-top), making it difficult to isolate object-only motion fidelity.

To address these limitations, we propose \metrics{} to explicitly isolate object-level motion by measuring how faithfully the foreground object moves from the input frame towards a user-defined target, while minimizing background and other foreground influence. We further introduce stable-camera, object-only benchmarks derived from DAVIS2017-test~\cite{DAVIS_Perazzi2016} and ObjMove-B~\cite{yu2025objectmovergenerativeobjectmovement}. Unlike datasets constructed by extracting trajectories from real videos using trackers (\eg, CoTracker~\cite{karaev23cotracker}), where trajectories are inherently entangled with camera motion~\cite{DAVIS_Perazzi2016,Huang2021,miao2022large}, our benchmarks decouple object dynamics from camera motion, enabling cleaner, fairer comparisons across image-to-video pipelines.
\vspace{-10pt}




\section{Methods}
\label{sec:method}

\ours{} is a training-free pipeline for interactive cut-and-paste object motion editing from a single image, designed around a practical constraint: users know \emph{where} they want an object to move, but not \emph{how}, they lack professional knowledge to design physically plausible trajectories, and have no 
access to depth maps, motion fields, or keyframes.

To operate under these constraints, \ours{} reformulates object motion editing as an FLF2V generation task. As illustrated in \cref{fig:pipeline}, the pipeline proceeds in three stages: (1) constructing a consistent target frame via semantic-guided object insertion and background inpainting; (2) synthesizing motion by conditioning the FLF2V generator on the first and target frames; and (3) selecting the best generation via \search{}, our early-step attention-consensus seed selection strategy. The pipeline and evaluation methods are detailed below.
\vspace{-10pt}


\subsection{Object Motion Editing as FLF2V}\label{sec:flf2v}

\subsubsection{Task Definition.}
Most prior motion-control interfaces assume expert-specified trajectories or dense guidance signals. In contrast, our goal is to reduce user burden and support non-expert users who provide only a coarse intent: the desired target location of the object. This minimal-input setting naturally motivates a formulation that operates from endpoint constraints rather than full trajectories.
To further decouple object motion from unintended camera dynamics, we consider a setting in which the object moves within a static background, so that the object motion is determined solely by the object transformation.  
\vspace{-10pt}

\subsubsection{FLF2V Reformulation.}
Recent advancements in FLF2V control for video generation~\cite{jiang2025vace} enables a model ${\cal M}$ to synthesize a video sequence ${\cal V}$ by blending between the given first frame ${\cal I}_f$ and last frame ${\cal I}_l$ conditions, guided by a text prompt ${\cal T}$. Motivated by the strong controllability offered by this paradigm, we reformulate object motion editing as an FLF2V generation task.
Specifically, given an input image serving as the first frame ${\cal I}_f$ and a mask of user-specified object of interest ${\cal O}_f$, we construct a synthesized last frame $\tilde{\cal I}_l$ that encodes the desired object transformation between the starting and ending states, denoted as ${\bf T}_{f\to l}$. The object mask can be readily obtained using click-then-segment models, such as SAM2~\cite{ravi2024sam2segmentimages}.
The video generation model ${\cal M}$ is then conditioned on $({\cal I}_f, \tilde{\cal I}_l)$ to produce a temporally coherent sequence in which the object moves from its initial location to the user-specified target location, guided either by a text prompt ${\cal T}$ provided by the user or an automatically generated prompt $\tilde{\cal T}$ from a visual-language model (VLM).
\vspace{-10pt}

\subsubsection{Semantic-guided Object Placement.}
To obtain a semantically reasonable target transformation ${\bf T}_{f\to l}$, we design a semantic-guided object placement pipeline that samples viable target locations. We query a VLM (\eg, Qwen2.5-VL~\cite{bai_qwen25-vl_2025}) to infer plausible placement regions conditioned on the scene context (\eg, ``on the road'', ``on the table'', ``in the sky'') and convert these textual guidance into spatial masks using SAM2~\cite{ravi2024sam2segmentimages}, as illustrated in \cref{fig:data_gen_example}. This provides users intuitive guidance on where an object can be reasonably placed, easing the user's effort in specifying target locations.
\vspace{-10pt}

\subsubsection{Synthesizing the Last-Frame Condition.}
While directly relocating objects with off-the-shelf image editing tools \cite{machalek-2020-kontext, wu2025qwenimagetechnicalreport} is a viable option, reliable and controllable object placement remains challenging for current open-source models. Such tools often introduce scene inconsistencies, such as geometric distortions, illumination mismatches, or unintended object replacements (\cref{fig:placement}).
Since our primary goal is to enable object motion editing through FLF2V, we instead synthesize the last frame $\tilde{\cal I}_l$ by generating a composite image with deterministic object placement. This ensures the object's ending state (including the location, rotation, and scale) is precisely encoded in the last frame.
To construct $\tilde{\cal I}_l$, we first adopt an inpainting model~\cite{zhao2025ObjectClear} to separate the foreground object (object of interest) from the background of the first frame ${\cal I}_f$, using the provided object mask ${\cal O}_f$. We then apply the affine transformation ${\bf T}_{f\to l}$ to the extracted object and paste the transformed instance to the inpainted background to form the composite image. A detailed analysis of different background inpainting methods is provided in the supplementary. Finally, we employ an image editing model~\cite{machalek-2020-kontext} to refine the composite image, producing a more natural and coherent final frame.
\vspace{-10pt}

\subsection{Robust Generation through Early-Step Attention Guidance}
\label{sec:aceseed}
\vspace{-5pt}
Current state-of-the-art video diffusion models (e.g., Wan) typically adopt a spatio-temporal VAE with an encoder-decoder pair that maps videos between pixel space and a compact latent space. A Diffusion Transformer (DiT) then performs flow matching in the latent space along a normalized time $t\in[0, 1]$ following a straight-line path
\begin{align}
    {\bf x}_t = (1 - t) {\bf x}_0 + t {\bf x}_1,
\end{align}
where ${\bf x}_0\sim {\cal N}({\bf 0}, {\bf I})$ is a Gaussian noise latent and ${\bf x}_1$ is the latent encoding of the conditioning input (e.g., first/last-frame conditions under FLF2V) produced by the VAE encoder.
In practice, we discretize $t$ into $T{+}1$ uniform grids, $s_i = i / T$ for $i=0,..., T$, yielding $T$ Euler updates with step size $\Delta t = 1/T$, i.e.,
\begin{align}
    {\bf x}_{s_{i+1}} = {\bf x}_{s_i} + \Delta t\cdot v_{\theta}({\bf x}_{s_i}, s_i, {\bf c}_{txt})
\end{align}
where the model predicts the velocity $v_{\theta}({\bf x}_t, t, {\bf c}_{txt})$ (trained to match ${\bf x}_1-{\bf x}_0$), conditioned on the text embedding ${\bf c}_{txt}$ of the prompt ${\cal T}$. After integrating to $t=1$, the resulting clean latent ${\bf x}_{s_T}\approx {\bf x}_1$ is then decoded into pixel space by the VAE decoder to obtain the generated video.

Since the initial random seed determines the noise latent ${\bf x}_0$ and thereby the flow path and velocity field estimate, the resulting video quality can vary substantially across seeds. More specifically, undesirable seeds can yield outlier trajectories, temporal artifacts, or degraded object fidelity. To mitigate this, we introduce a trajectory-preview-driven seed selection strategy, \search{}, that filters unstable seeds before full generation.
\vspace{-10pt}

\subsubsection{Early-step Trajectory Preview.}
In FLF2V, the conditioning frames ${\cal I}_f$ and ${\cal I}_l$ anchor the endpoints of the object's motion, constraining the latent subspace the model can explore when generating trajectories. This structure allows us to exploit the DiT attention maps to preview how the model intends to connect the two states.
We observe that modern video diffusion models establish coarse motion structure early in the denoising process~\cite{hong2022cogvideo, HaCohen2024LTXVideo, wan2025, jiang2025vace}.
During the first $t_{\text{early}}$ denoising steps (\eg, $\leq 10$ out of $T{=}50$), we extract self-attention maps from a fixed subset of DiT attention blocks and aggregate them over object-relevant tokens derived from the first-frame mask ${\cal O}_f$.
These early-step attention maps already reveal emerging object dynamics, provide interpretable previews of candidate motion behaviors, and enable users to adjust the target placement (or choose among candidates) before committing to the full video synthesis.
\vspace{-10pt}

\subsubsection{Attention-Consensus Seed Selection (\search{}).}
While trajectory previews allow users to intercept video generation early, they also offer a useful signal for {\bf automatic seed selection}.
Given a set of candidate noise latents $\{{\bf x}_0^{(i)}\}_{i=1}^{N}$, we roll out each seed for $t_{\text{early}}$ denoising steps and extract a sequence of attention representations from the $m$-th to the $n$-th DiT layer:
\begin{equation}
{\cal A}^{(i)}=\{{\bf A}^{(i)}_m,\dots,{\bf A}^{(i)}_n\}.
\end{equation}

To capture both object movement and scene-level stability, we aggregate attention using tokens corresponding to the \emph{inverse} of the first-frame object mask ${\cal O}_f$, i.e., the non-object region. This aggregation yields an early-step attention signature ${\bf h}_i$ that reflects global dynamics (\eg, background drift) while 
remaining sensitive to object relocation through its contextual interactions with the surrounding scene.

We treat consensus as a robustness prior, and hypothesize that seeds whose early-step attention patterns align most strongly with the group are least likely to produce outlier trajectories. Accordingly, we compute a consensus score $a^{(i)}$
that measures how well ${\bf h}_i$ agrees with other signatures, and select the seed via
\begin{equation}
k=\arg\max_i a^{(i)}\quad\quad a^{(i)}=sim({\bf h}_i, {\bf \bar h}_{j\neq i})
\end{equation}
where ${\bf \bar h}_{j\neq i}$ denotes the mean vector of the other candidates, and $sim(\cdot, \cdot)$ is a similarity measure. The full procedure is summarized in \cref{alg:seed}.
Details of the similarity metrics and corresponding ablations are provided in \cref{sec:experiment}.
Finally, the selected seed proceeds to full sampling.
\vspace{-5pt}

\begin{algorithm}[t]
\caption{\search{}: Attention Consensus for Early-step Seed Selection}
\label{alg:seed}
\begin{algorithmic}
\REQUIRE Number of seeds $N$, early step $t_{\text{early}}$
\ENSURE Selected seed index $k$
\FOR{$i = 1$ to $N$}
    \STATE Roll out seed ${\bf x}_0^{(i)}$ for $t_{\text{early}}$ steps
    \STATE Extract self-attention maps ${\cal A}^{(i)}$ at $t_{\text{early}}$
    \STATE Aggregate attention over object-relevant tokens to obtain ${\bf h}_i$
\ENDFOR
\FOR{$i=1$ to $N$}
    \STATE Compute similarity score $a^{(i)}=sim({\bf h}_i, {\bf \bar h}_{j\neq i})$
\ENDFOR
\STATE $k \gets \arg\max_i a^{(i)}$
\STATE \RETURN $k$
\end{algorithmic}
\end{algorithm}

\subsection{Benchmarks and Object-Level Motion Metrics}
\label{sec:benchmark_metric}

Existing video benchmarks and whole-frame metrics often entangle foreground motion with camera/background changes, making them insensitive to artifact-prone object trajectories (\cref{fig:traj}). We therefore introduce {\bf stable-camera, object-only motion benchmarks}, \davis{} and \objmove{}, and propose {\bf object-level motion metrics}, \metrics{}, that isolate object fidelity by measuring how faithfully the edited object moves toward a user-defined target, while minimizing the influence of the background and other foreground elements. These benchmarks and metrics enable clean evaluation of object-only motion controllability and are used throughout our experiments (\cref{sec:experiment}).
\vspace{-10pt}

\subsubsection{{\bf\ours{}} Benchmarks.}
\label{sec:s2m_data}
We introduce \ours{} benchmarks consisting of first-last frame pairs under a static-background setting. An example is shown in \cref{fig:traj}-bottom. Given a first-frame image ${\cal I}_f$ and its object mask ${\cal O}_f$, we synthesize the last frame ${\cal \tilde I}_l$ using the compositing method introduced in \cref{sec:flf2v}.
We construct the \davis{} benchmark by applying the semantic-guided object placement pipeline (\cref{sec:flf2v}) to a subset of DAVIS2017-test~\cite{DAVIS_Perazzi2016}. 
Given the placement mask produced by the pipeline, we sample a target position along with scale and rotation factors drawn from preset ranges. Additional details and examples are included in the supplementary material.
Because many objects in the DAVIS dataset do not exhibit self-locomotion, this process yields 20 first-last frame pairs across 10 scenarios. Similarly, we create the \objmove{} benchmark by generating last-frame composite images for 38 images from the ObjMove-B dataset~\cite{yu2025objectmovergenerativeobjectmovement}, utilizing the provided source and target masks.
\vspace{-10pt}


\subsubsection{\metrics{} Metrics.}\label{sec:metrics}
\begin{figure}[t!]
    \centering
    \begin{minipage}[h]{.615\linewidth}
        \vspace{8pt}
        \includegraphics[width=\linewidth]{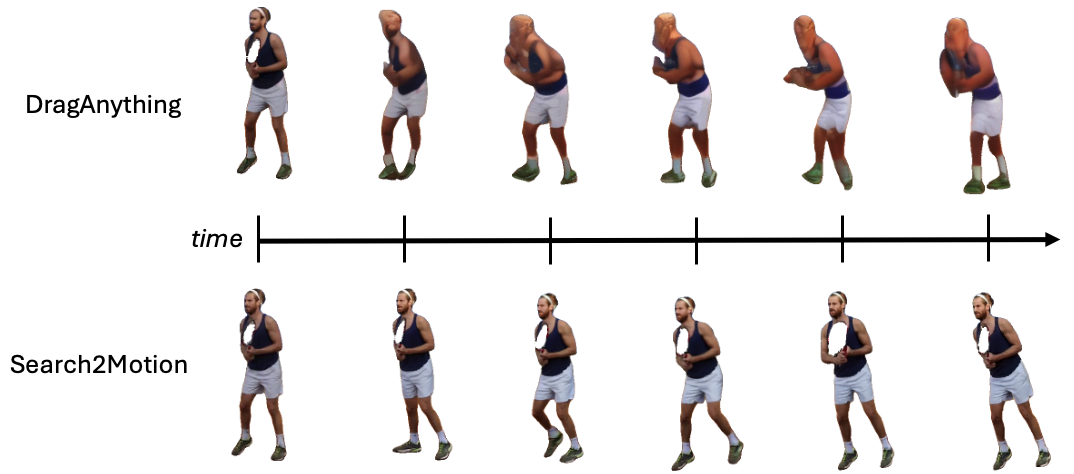}
        \caption{FLF2V-obj metrics provides object-centric insight by isolating the object from the scene and evaluating object consistency across the generated sequence. 
        \ours{} produces high-fidelity object movement and maintains object consistency across the generated sequence compared to DragAnything.}
        \label{fig:tennis-vest_1}
    \end{minipage}
    \hfill
    \begin{minipage}[h]{.37\linewidth}
        \includegraphics[width=\linewidth]{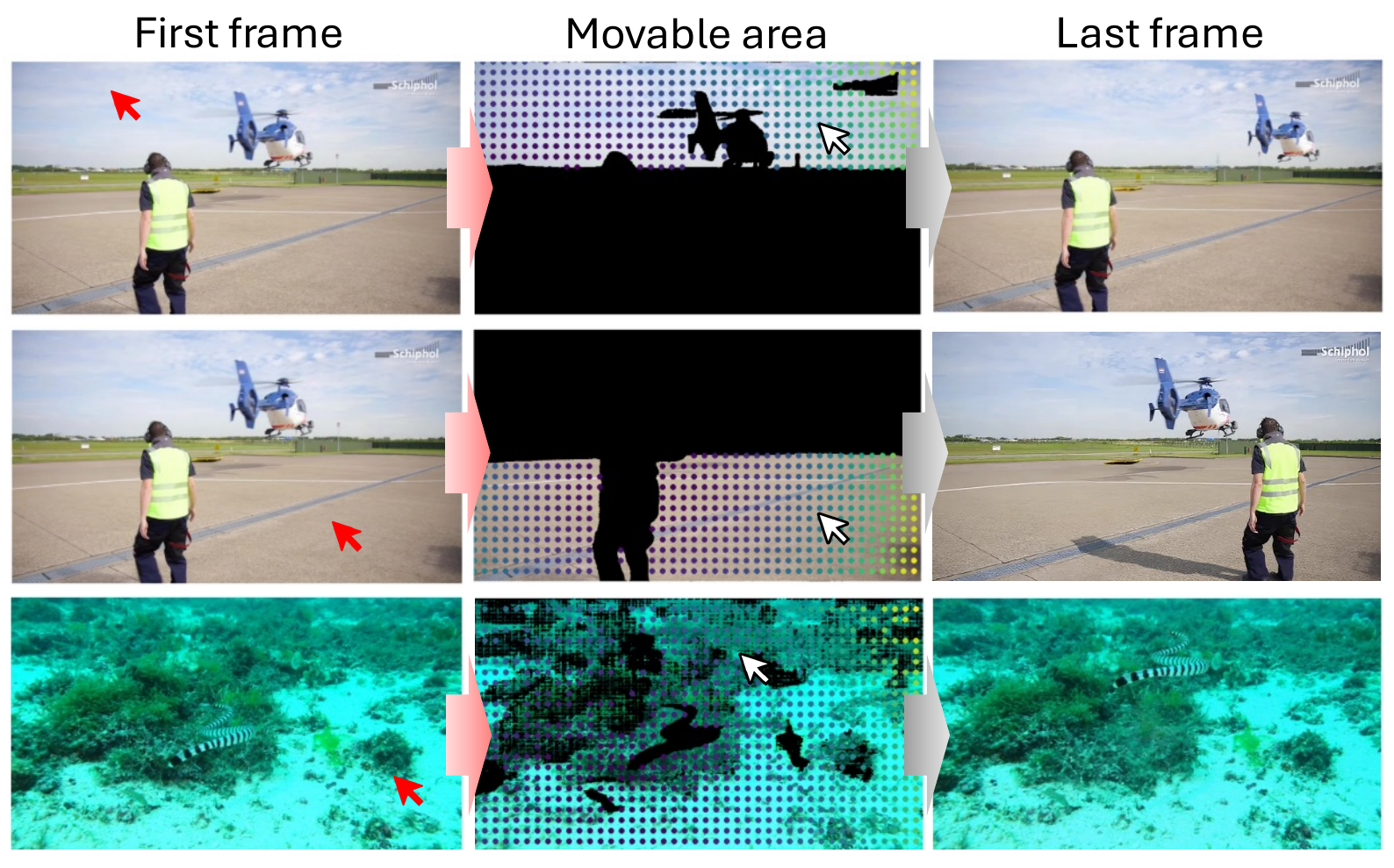}
        \vspace{-20pt}
        \captionof{figure}{
        The VLM identifies and points (red) to the plausible placement area in the first frame (\eg, the sky) for the target object (\eg, a helicopter). SAM2 segments the placement mask to define the movable area. The user (white) can then select the placement location accordingly to construct the last frame.
        }
        \label{fig:data_gen_example}
    \end{minipage}
\end{figure}

Existing video-generation metrics (\eg VBench~\cite{huang2024vbench}) evaluate whole-frame fidelity, inevitably coupling foreground and background artifacts. To provide a complementary evaluation, we introduce \metrics{}, a set of {\bf object-centric metrics} that explicitly quantify object fidelity by isolating the target object from the surrounding scene.
An example sequence of masked objects is shown in \cref{fig:tennis-vest_1}, illustrating the benefit of evaluating objects independently.
We propagate the first-frame object mask throughout the generated video using SAM2 \cite{ravi2024sam2segmentimages}, obtaining an object mask for each frame at a resolution of $224\times224$.
We then compute a set of {\it object consistency} metrics based on LPIPS\cite{zhang2018unreasonableeffectivenessdeepfeatures} distance and DINOv2\cite{oquab2024dinov2learningrobustvisual} cosine similarity for two different schemes respectively: (i) between consecutive object crops, and (ii) between the first-frame object crop and all other frames.
This yields four metrics that capture the coherent appearance and temporal stability of the edited object across the generated sequence. These metrics are reference-free, as they do not rely on ground-truth object trajectories, but instead assess the internal consistency of the generated object motion. We show that \metrics{} are well aligned to human preference in \cref{tab:human_study}.
\vspace{-10pt}




\section{Experiments}\label{sec:experiment}

\subsection{Experimental Setup}

The experiments are designed to evaluate three components: (i) the effectiveness of our FLF2V formulation for object motion editing, (ii) the robustness of \search{} for seed selection via early-step attention consensus, and (iii) the utility of \metrics{} for isolating and assessing object-level fidelity.
We evaluate \ours{} using two FLF2V video generation models (i.e., VACE-1.3B and Wan2.2-5B) on two stable-camera, object-only motion benchmarks, \davis{} and \objmove{} (introduced in \cref{sec:s2m_data}).
For seed selection, to ensure a fair comparison, we use the same set of random seeds ($N=10$) with $t_{early}=10$ and $T=50$ denoising steps, unless specified otherwise. 
We report VBench~\cite{huang2024vbench} metrics -- including {\it subject consistency}, {\it background consistency}, {\it temporal flickering}, {\it motion smoothness}, {\it aesthetic quality}, and {\it imaging quality} (all of which support evaluation with custom videos and prompts) -- for overall video quality assessment, and our proposed \metrics{} metrics for detailed object-level fidelity analysis.
\vspace{-10pt}

\begin{table}[t!]
    \centering
    \captionof{table}{Comparison to existing motion-control methods on VBench metrics.}\label{tab:vbench}
    \vspace{-5pt}
    \setlength{\tabcolsep}{4pt}
    \resizebox{\linewidth}{!}{
    \begin{tabular}{llcccccc}
    \toprule
    Dataset & Method & Subject & Background & Temporal & Motion & Aesthetic & Imaging \\
    & & Consistency & Consistency & Flickering & Smoothness & Quality & Quality \\
    \midrule
    \objmove{} & DragAnything~\cite{wu2024draganything} & 89.73	& 92.13 & 95.15	& 97.62 &	55.06 &	57.94 \\
     & TTM - Wan2.2-5B~\cite{singer2025ttm} & 93.92 & 95.29 & 98.00 & 98.59 & 59.42 & 70.03 \\ \cline{2-8}
    & \ours{} - VACE-1.3B & 94.77 & 95.98 &	97.99 &	98.75 &	{\bf 61.96} & 71.34 \\ 
    & \ours{} - Wan2.2-5B & {\bf 95.19} & {\bf 96.07} &	{\bf 99.00} & {\bf 99.45} &	56.86 &	{\bf 71.54} \\
    \midrule
    \davis{} & DragAnything~\cite{wu2024draganything} & 87.11 & 88.77 & 95.51 & 97.75 & 42.34 & 53.46 \\
    & TTM - Wan2.2-5B~\cite{singer2025ttm} & 96.26 & 96.09 & 98.45 & 98.96 & 46.69 & 67.36 \\ \cline{2-8}
    & \ours{} - VACE-1.3B & 95.69 & 95.89 & 98.53 & 99.00 & {\bf 48.00} & {\bf 69.58} \\ 
    & \ours{} - Wan2.2-5B & {\bf 96.94}	& {\bf 96.52} & {\bf 99.34} & {\bf 99.57} & 47.04 & 69.50 \\
    \bottomrule
    \end{tabular}
}
    \vspace{5pt}
    
    \centering
    \captionof{table}{Comparison to existing motion-control methods on \metrics{} metrics.}\label{tab:flf2v}
    \vspace{-5pt}
    \setlength{\tabcolsep}{8pt}
    \resizebox{\linewidth}{!}{
    \begin{tabular}{llcccc}
    \toprule
    Dataset & Method  & DINOv2$\uparrow$ & DINOv2$\uparrow$ &LPIPS$\downarrow$ & LPIPS$\downarrow$ \\
    &&& first frame &  & first frame \\
    \midrule
    \objmove{} & DragAnything~\cite{wu2024draganything} & 90.74 & 75.12 & .0991 & .2213 \\
    & TTM - Wan2.2-5B~\cite{singer2025ttm} & 96.44 & 80.26 & .0468 & .2384 \\
    \cline{2-6}
    & \ours{} - VACE-1.3B & 96.92 & \textbf{84.46} & .0498 & \textbf{.1965} \\
    & \ours{} - Wan2.2-5B & \textbf{97.11} & 84.31 & \textbf{.0416} & .2074 \\
    \midrule
    \davis{} & DragAnything~\cite{wu2024draganything} & 83.79 & 68.23 & .1671 & .2751 \\
    & TTM (Wan2.2-5B)~\cite{singer2025ttm} & 95.25 & 78.81 & .0656 & \textbf{.2464} \\
    \cline{2-6}
    & \ours{} - VACE-1.3B & 95.08 & 80.90 & .0804 & .2653 \\
    & \ours{} - Wan2.2-5B & \textbf{97.55} & \textbf{89.20} & \textbf{.0557} & .2556 \\
    \bottomrule
    \end{tabular}
}
    \vspace{-15pt}
\end{table}

\begin{table}[t!]
    \centering
    \caption{Object center distance (ObjMC) and center-aligned object mask IoU (CA-IoU) between the penultimate frame in the generated video and the given target frame.}\label{tab:obj_end_state}
    \vspace{-5pt}
    \setlength{\tabcolsep}{8pt}
    \resizebox{.7\linewidth}{!}{
    \begin{tabular}{llcc}
    \toprule
    Dataset & Method & ObjMC$\downarrow$ & CA-IoU$\uparrow$ \\
    \midrule
    \objmove{} & DragAnything~\cite{wu2024draganything} & 58.12 & 0.5619 \\
     & TTM - Wan2.2-5B~\cite{singer2025ttm} & 175.02 & 0.4460 \\ \cline{2-4}
     & \ours{} - Wan2.2-5B & {\bf 9.96} & {\bf 0.8818} \\ \midrule
    \davis{} & DragAnything~\cite{wu2024draganything} & 36.61 & 0.4290 \\
     & TTM - Wan2.2-5B~\cite{singer2025ttm} & 105.79 & 0.3686 \\ \cline{2-4}
     & \ours{} - Wan2.2-5B & {\bf 3.72} & {\bf 0.8211} \\
    \bottomrule
    \end{tabular}
}
\vspace{5pt}
\end{table}



\subsection{Comparison to Trajectory-Based Control Baselines}
\label{sec:exp_main}

We compare \ours{} against trajectory-based control methods under the same object masks and displacement settings.
DragAnything~\cite{wu2024draganything} injects motion control by smoothing a user-defined trajectory, while TTM~\cite{singer2025ttm} uses a warped reference video synthesized by pasting the object crop along a provided trajectory path. In contrast, our setting assumes that users specify only the target object transformation, not the full trajectory. For fair comparison, we linearly interpolate between the initial and the final object positions to construct the input trajectories required by baselines.

\begin{figure}[t!]
    \centering
    \includegraphics[width=.85\linewidth]{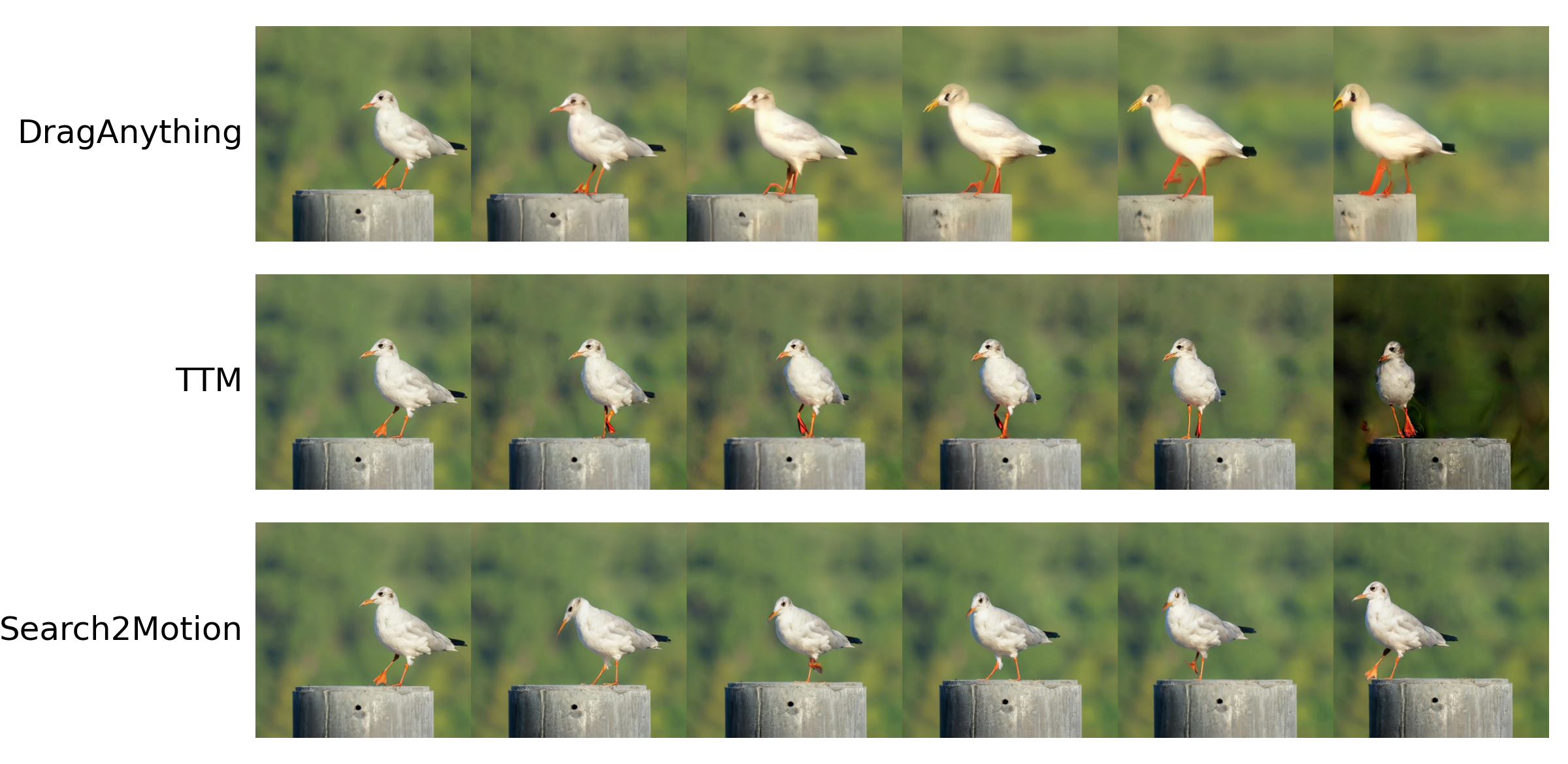} \\\vspace{-5pt}
    \includegraphics[width=.85\linewidth]{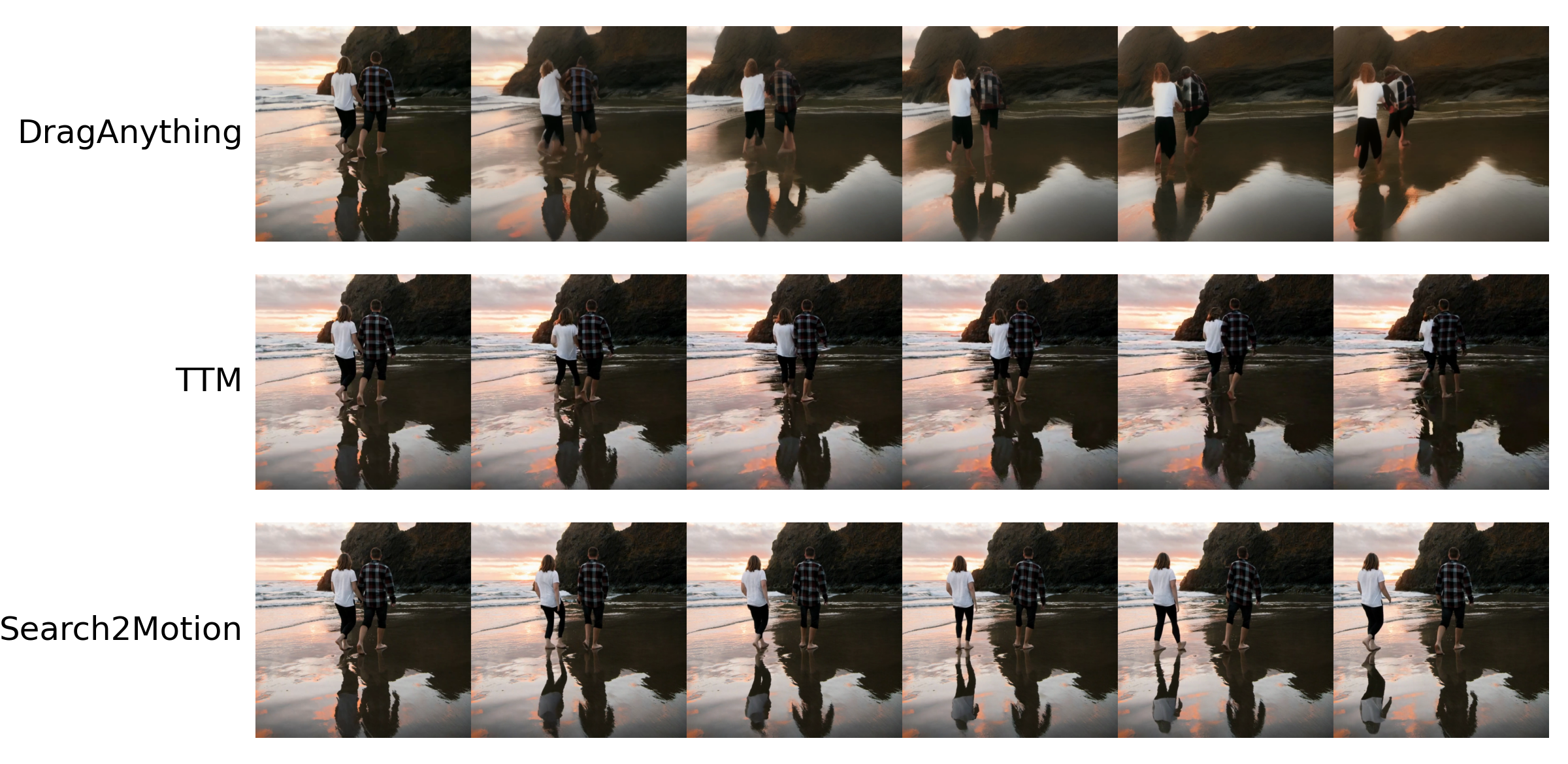} \\\vspace{-5pt}
    \includegraphics[width=.85\linewidth]{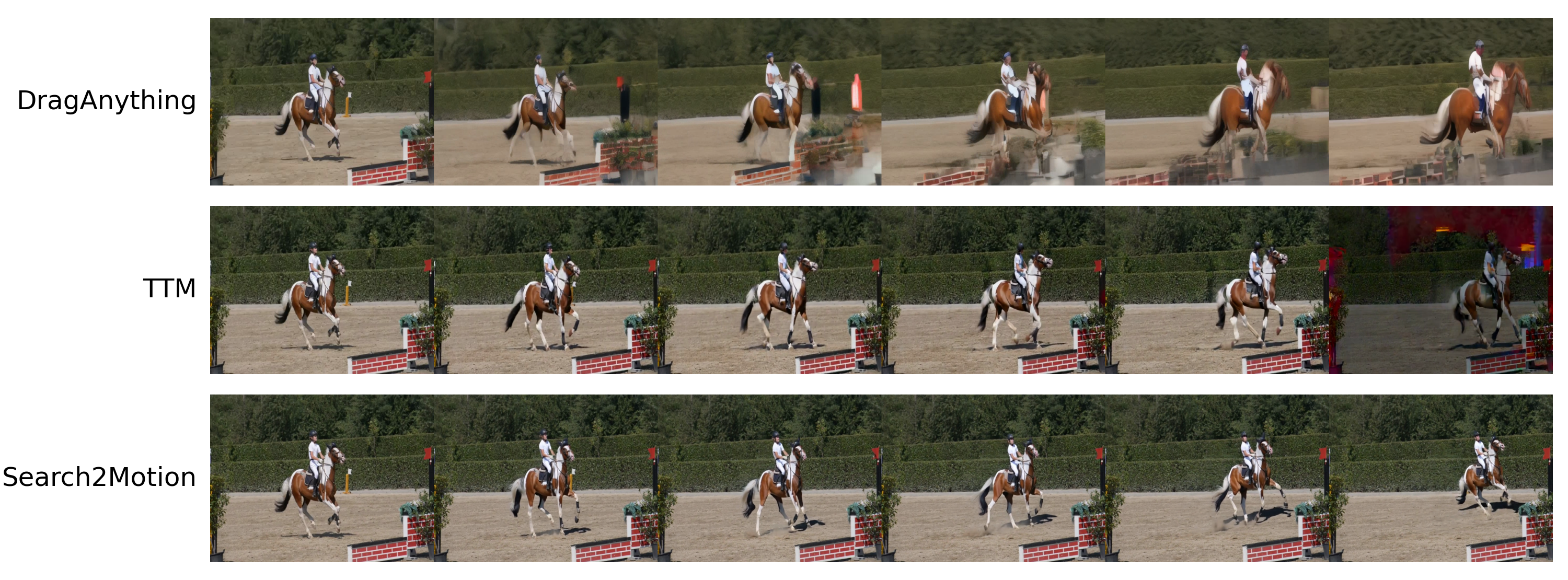} \\\vspace{-5pt}
    \captionof{figure}{Qualitative results of \ours{} video generation comparing against DragAnything and TTM on \objmove{} (top two) and \davis{} (bottom).}
    \label{fig:video_example}
\end{figure}

\cref{tab:vbench} and \cref{tab:flf2v} present the VBench and \metrics{} results, respectively.
Overall, \ours{} generates higher-quality videos across both benchmarks, with particularly strong improvements in object-level fidelity as captured by \metrics{}. This is also visible in the qualitative examples shown in \cref{fig:video_example}, where images are uniformly sampled from the generated videos for visualization.
DragAnything frequently exhibits spatial drift, inconsistent boundary blending, or scene-object incompatibility for larger moves.  
In contrast, \ours{} and TTM generate more stable transitions, with fewer artifact-prone trajectories. However, TTM often fails to maintain accurate target placement and visual quality in later frames, producing objects at unintended locations or with incorrect poses.
To quantify this behavior, we further evaluate the object's state in the penultimate frame\footnote{We exclude the final frame because FLF2V models typically copy the last-frame condition over to the output, which may be unfair to the compared baselines.} of the generated video and compare it to the given last-frame target state.
Using SAM2 to propagate the object mask through the generated sequence, we compute ObjMC, the center distance between the predicted and target object locations. To isolate pose similarity from placement accuracy, we evaluate the IoU between the predicted and target object masks after aligning their centers (visualization shown in the supplementary material), denoted by CA-IoU.
Results in \cref{tab:obj_end_state} show that \search{} achieves substantially better control of target object location and pose compared to the baselines, consistent with our visual observations. 

\begin{figure}[t]
    \centering
    \begin{minipage}[h]{0.51\linewidth}
        \vspace{2pt}
        \includegraphics[width=\linewidth]{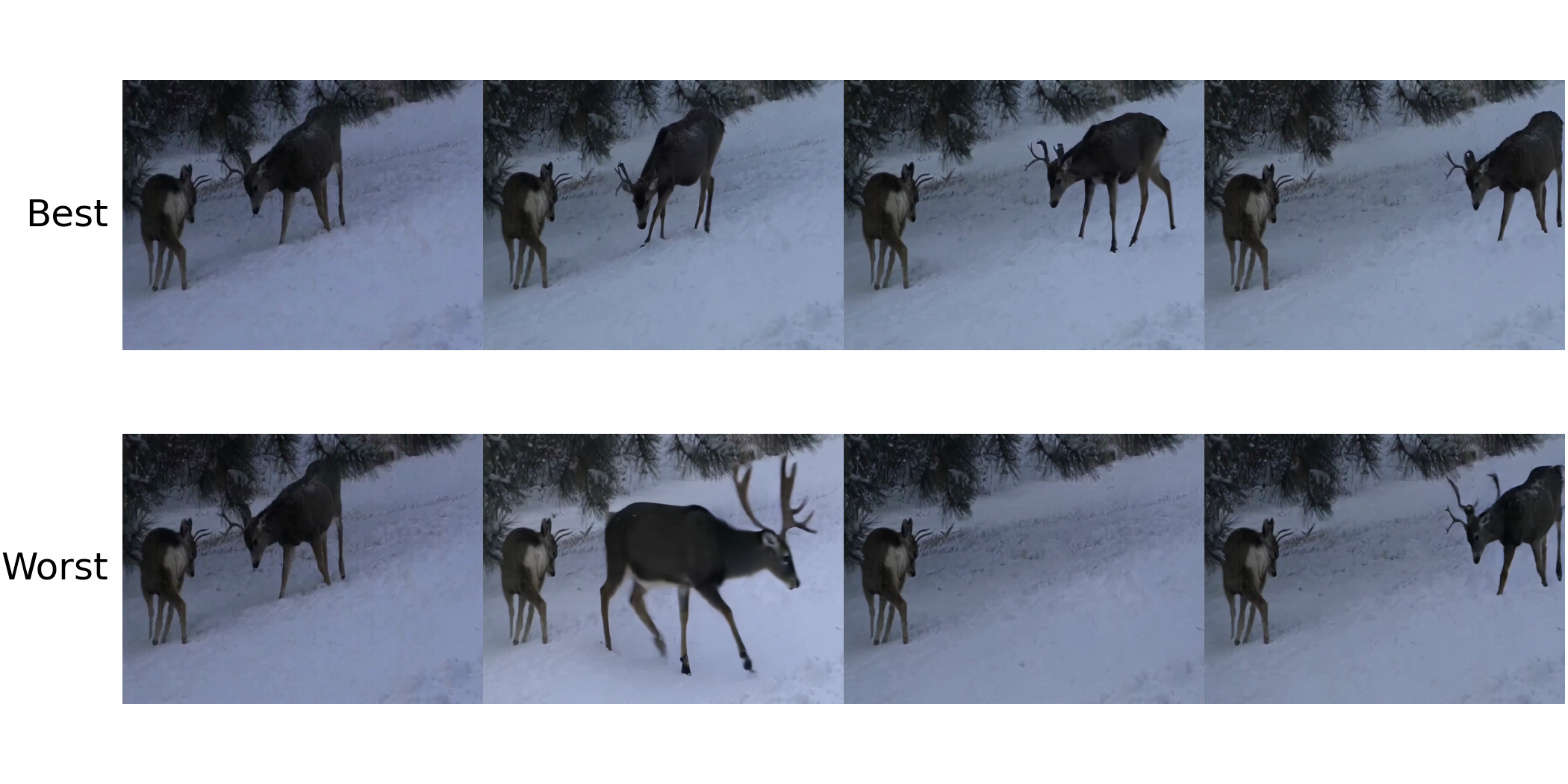}
    \end{minipage}
    \hfill
    \begin{minipage}[h]{0.48\linewidth}
        \includegraphics[width=\linewidth]{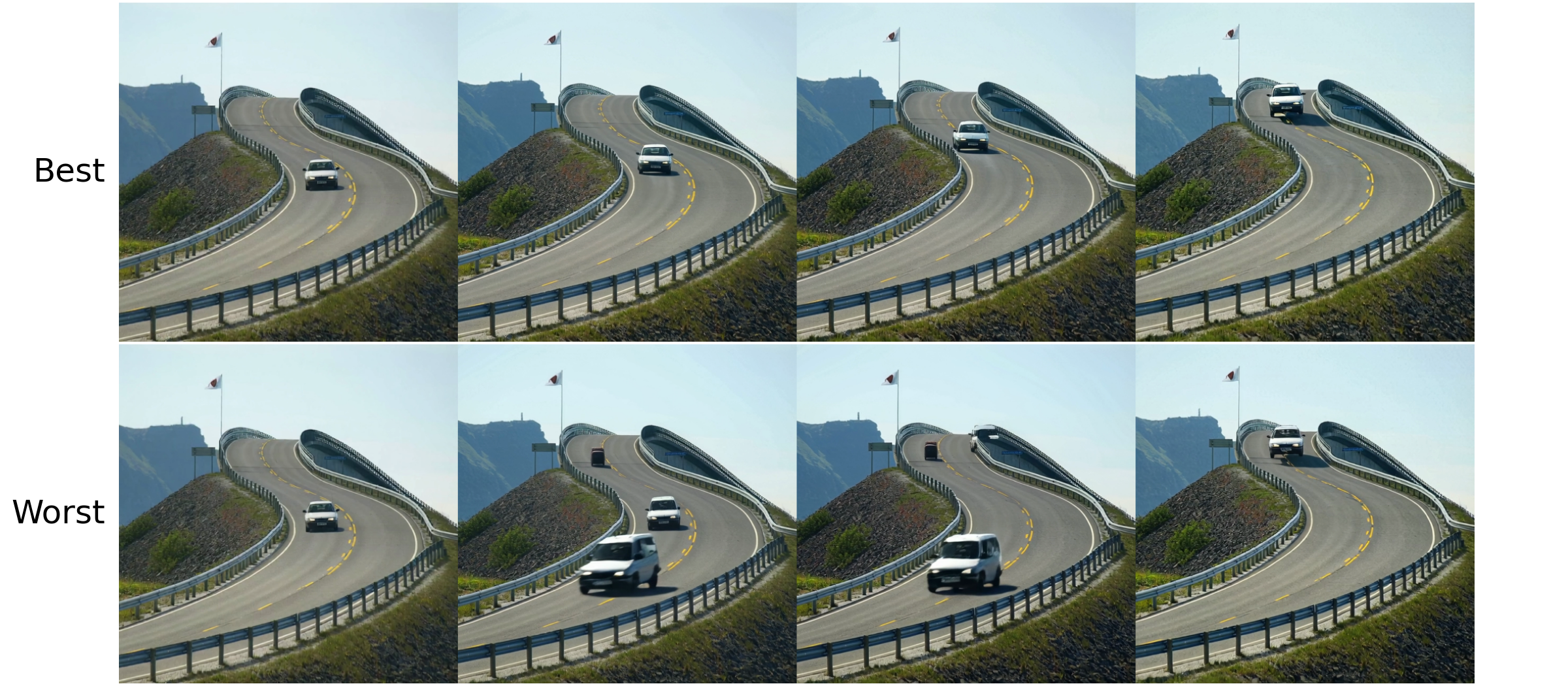}
    \end{minipage}
    \vspace{-15pt}
    \captionof{figure}{\textbf{Best-Worst comparisons 
    ranked by \search{} algorithm.} Our noise-seed search via early-step attention consensus under the optimal setup, reliably identifies the highest-quality and lowest-quality samples, with worst cases showing strong motion artifacts. Top: deer (\davis{}). Bottom: car (\objmove{}). Additional qualitative comparisons are provided in the supplementary material.}
    \label{fig:worst_vs_best}
    \vspace{-5pt}
    \vspace{5pt}
    \centering
    \begin{minipage}[h]{\linewidth} %
        \centering
         \includegraphics[width=0.8\linewidth]{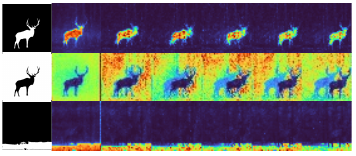}
         \vspace{-10pt}
        \captionof{figure}{Early-stage self-attention visualization under different token-masking strategies. For each sample, from top to bottom: attention computed with the \textbf{foreground (fg) object mask}, \textbf{inverse foreground mask (bg)}, \textbf{placement mask (pl)}. It shows that bg can provide the most rich information (comparing to fg) with a relatively more consistent area coverage among different samples (comparing to pl). Additional visualization is shown in the supplementary.}
    \label{fig:mask_ablation}
    \end{minipage}
\end{figure}

\vspace{-10pt}
\subsection{Ablation on Seed Selection Strategies}
\label{sec:exp_aceseed}
\vspace{-5pt}
\search{} is an inference-time seed selector using early-step attention consensus. We conduct extensive ablations on its components as follows.

\vspace{-10pt}
\subsubsection{Effectiveness of Seed Selection.}
Using the same set of $N$ candidate seeds, we first evaluate the efficacy of \search{} by comparing its best-ranked seed against the average performance of the non-selected seeds (none). 
As shown in \cref{tab:search_wanfun}, \search{} consistently improves both VBench and \metrics{} metrics across datasets, indicating reduced sensitivity to unlucky initializations. Results for VACE-1.3B show a similar trend and are included in the supplementary.
We also compare with BANSA~\cite{kim2025modelknowsbestnoise}, a concurrent seed-selection method based on model uncertainty. Since the official code is unavailable, we implement it according to their paper.
In our setting, BANSA
generally underperforms \search{}.

\vspace{-10pt}
\subsubsection{Best--Worst Separation.}
\cref{fig:worst_vs_best} visualizes samples ranked by \search{}, showing that seeds with low consensus scores tend to produce obvious motion artifacts and unstable dynamics, whereas high-consensus seeds yield cleaner object motion and better scene coherence. This qualitative separation supports our claim that early-step attention consensus provides a reliable, model-internal signal for avoiding outliers without look-ahead sampling or external reward models.

\begin{table}[t!]
    \centering
    \captionof{table}{Seed selection ablation on VBench and \metrics{} metrics using Wan2.2-5B. $\dagger$ denotes the average performance of non-selected seeds.}
    \label{tab:search_wanfun}
    \vspace{-5pt}
    \setlength{\tabcolsep}{3pt}
    \resizebox{\linewidth}{!}{
    \begin{tabular}{ll|cccc|cccc}
    \toprule
    Dataset & Seed Selection & Subject$\uparrow$ & Background$\uparrow$ & Temporal$\uparrow$ & Motion$\uparrow$ & DINOv2$\uparrow$ & DINOv2$\uparrow$ & LPIPS$\downarrow$ & LPIPS$\downarrow$ \\
    & & Consistency & Consistency & Flickering & Smoothness & & first frame &  & first frame \\
    \midrule
    \objmove{} & none$\dagger$ & 95.14 & 95.88 & 98.68 & 99.30 & 96.99 & 82.60 & .0435 & .2207 \\   
    & BANSA~\cite{kim2025modelknowsbestnoise} & 94.81 & 95.49 & 98.36 & 99.25 & 97.03 & 83.02 & .0446 & .2177 \\ \cline{2-10}
    & \search{} & {\bf 95.19} & {\bf 96.07} &	{\bf 99.00} & {\bf 99.45} & {\bf 97.09} & {\bf 84.31} & {\bf .0419} & {\bf .2075} \\
    \midrule
    \davis{} & none$\dagger$ & 96.45 & {\bf 96.52} & 99.30 & {\bf 99.57} & 95.70 & 78.31 & .0637 & .2739 \\
    & BANSA~\cite{kim2025modelknowsbestnoise} & 95.85 & 96.10 & 98.85 & 99.45 &	95.87 & 77.87 & .0568 & .2684 \\  \cline{2-10}
    & \search{} & {\bf 96.94} & {\bf 96.52} & {\bf 99.34} & {\bf 99.57} & {\bf 96.42} & {\bf 80.43} & {\bf .0557} & {\bf .2556} \\
    \bottomrule
    \end{tabular}
}
\end{table}


\vspace{-10pt}
\subsubsection{Number of Seeds $N$.}
Larger $N$ improves robustness but exhibits diminishing returns beyond $\sim$13 to 15 seeds. Results are shown in the supplementary.
\vspace{-10pt}

\subsubsection{Attention Layer Selection.}
Mid-to-deep layers (\eg, blocks 22--26 of 30) correlate best with downstream motion fidelity, while early/very deep layers are less predictive. Please refer to the supplementary.

\subsubsection{Similarity Metric.}
We consider cosine distance and Sinkhorn distance~\cite{sinkhorn_NIPS2013_af21d0c9}. The former measures the angular difference between two vectors, while the latter is a regularized variant of the Earth Mover's Distance (EMD), where we treat the Euclidean distance between pixel locations as the transport cost, enabling spatially aware comparisons between distributions. Both metrics yield similar results with consistent seed ranking under \metrics{} scores. For simplicity, we adopt cosine similarity metric as the distance measure in all our experiments.
\vspace{-10pt}
\subsubsection{Token Selection (fg vs bg vs placement).}
As shown in \cref{fig:mask_ablation}, aggregating attention using the \textbf{inverse foreground mask (bg)} yields the most reliable ranking under \metrics{}. We hypothesize bg tokens better capture both (i) global scene stability (\eg, background drift) and (ii) object--scene interactions induced by the relocation, whereas fg-only aggregation can miss relative motion when object attention remains spatially concentrated. The placement mask can capture both regions but varies substantially in area across samples, reducing stability. Therefore, we adopt bg aggregation for all reported \search{} results.





\vspace{-5pt}
\subsection{Human Preference Study}
\vspace{-10pt}

Given a first-last frame image pair $({\cal I}_f, {\cal I}_l)$, we sample $N=10$ random seeds and use \search{} to select the top-ranked seed. To assess how well \search{}, \metrics{}, and VBench align with human preferences, we generate full videos for all seeds and present them to human annotators.
For each pairwise comparison, each annotator is shown two videos corresponding to two different noise seeds (presented in a random order), and asked to choose the one that appears more realistic. With 10 seeds, this yields 45 unique pairs.
After collecting all pairwise judgments, each seed receives a count of how many times it was preferred, from which we derive a human-based ranking of all seeds.
We conduct this study for all samples in \davis{} with \ours{} using VACE-1.3B.

For each $({\cal I}_f, {\cal I}_l)$ pair, we then rank the same 10 seeds using \search{}, \metrics{}, and VBench, respectively.
To quantify alignment with human rankings, we report {\bf R$@$top-5} and {\bf R$@$botom-5} metrics, defined as the recall of the top-5 and bottom-5 human-ranked seeds, respectively, averaged over all $({\cal I}_f, {\cal I}_l)$ pairs.
\cref{tab:human_study} reports the result of this study.
All metrics achieve better alignment than {\it chance} (random ranking), where both R$@$top-5 and R$@$botom-5 equal 50.
Despite accessing only early-step attention maps without rolling out for the full video generation, \search{} achieves competitive alignment to human preference relative to VBench, which operates on RGB videos.
Moreover, \metrics{} exhibits stronger alignment than VBench, consistent with its design for object-level fidelity assessment.






\begin{table}[t!]
    \centering
    \caption{Recall scores using human ranking as relevant items from \davis{}. \metrics{} aligns best to human preference, highlighting its value as a set of new metrics. * denotes average recall over respective set of metrics.}\label{tab:human_study}
    \vspace{-5pt}
    \setlength{\tabcolsep}{14pt}
    \resizebox{.6\linewidth}{!}{
    \begin{tabular}{lcc}
    \toprule
    Ranking & R$@$top-5 & R$@$bottom-5 \\
    \midrule
    \search{} & 62.50 & 54.16 \\ \midrule
    VBench & 63.88* & 61.11* \\ 
    \metrics{} & \textbf{70.83}* & \textbf{78.12}* \\
    \bottomrule
    \end{tabular}
    }
\end{table}

\vspace{-10pt}

\section{Conclusion}
\vspace{-5pt}

\ours{} demonstrates controllable object-level motion in video generation does not require learning new priors or retraining. The necessary motion knowledge already exists within pretrained FLF2V models, and the central challenge is eliciting it reliably. This reframing has a practical implication: as video foundation models improve, Search2Motion inherits those gains for free, without re-engineering or fine-tuning.

A key finding is that early-step attention maps are predictive of downstream motion quality, but not uniformly so. Only a specific subset of mid-to-deep DiT attention blocks produces visualizations that spatially correspond to the coarse motion structure of the final generated video. This layer-specificity suggests that trajectory 
commitment is not a property of the denoising process as a whole, but is localized to particular layers in the network. ACE-Seed exploits this by treating consensus across seeds at these layers as a robustness prior, providing a cheap, model-internal 
quality signal without any look-ahead sampling. This insight generalizes beyond our pipeline: any frame-conditioned diffusion system may benefit from selective layer attention as an early indicator of generation quality.


Finally, our benchmark study reveals a systemic gap in how object motion is evaluated. Existing metrics reward whole-frame consistency, which can mask object-level failures entirely. We hope S2M-DAVIS, S2M-OMB, and FLF2V-obj encourage the community to evaluate object and camera dynamics separately, enabling fairer and more diagnostic comparisons across pipelines.

\section*{Acknowledgements}
We thank Sameer Sheorey, Kyle Min, Diana Wofk, Benjamin Ummenhofer, and Michael Paulitsch for their valuable feedback on this work.
\par\vfill\par
\clearpage  

%
%
\bibliographystyle{splncs04}
\bibliography{main}

\clearpage


\begin{center}
    {\Large\bfseries Supplementary Materials\par}
\end{center}
\appendix
\section{Methods}
This section provides additional implementation details for two components of the \ours{} pipeline omitted from the main paper for brevity: the background inpainting module selection (\cref{sec:supp_inpainting}) and the automated pipeline for benchmark dataset generation (\cref{sec:supp_benchmark}).

\subsection{Background Inpainting Module}
\label{sec:supp_inpainting}

\begin{figure}[b]
    \centering
    \includegraphics[width=\textwidth]{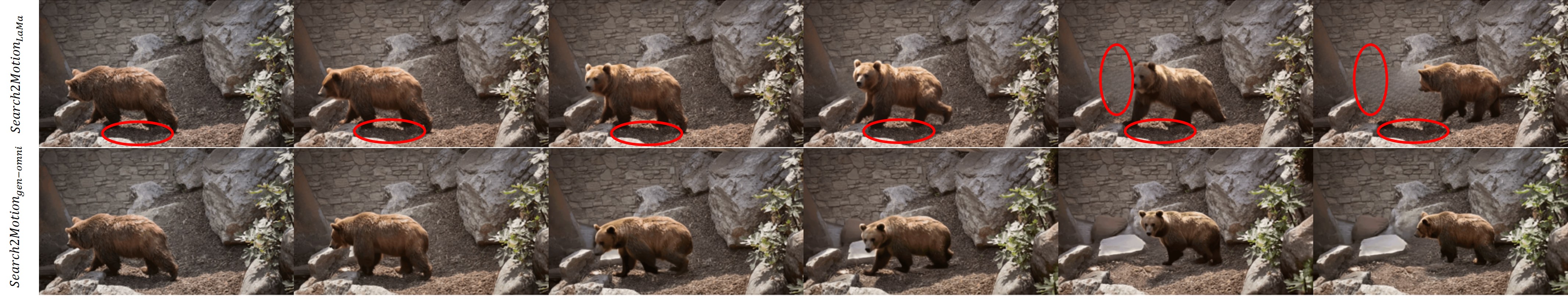}
    \vspace{-10pt}
    \caption{Comparison of our \ours{} using LaMa\cite{suvorov2021resolution} inpainting versus Generative Omnimatte \cite{lee2025generative-omnimatte}. With LaMa, last-frame's inpainting artifacts persist throughout the generated sequence, whereas our earlier method utilizing Generative Omnimatte produces cleaner backgrounds and yields overall higher video quality.}
    \label{fig:inpaint_artifact}
\end{figure}

A reliable background inpainting module is critical to the Search2Motion pipeline. Since the inpainted background serves as the foundation for target frame construction, any artifacts (residual object boundaries, inconsistent shadows, or texture discontinuities) propagate directly into the FLF2V generator and persist throughout the generated video (\cref{fig:inpaint_artifact}). This makes inpainting quality a bottleneck for downstream motion fidelity, motivating a careful selection of the inpainting component.

\begin{figure}[t!]
    \centering
    \includegraphics[width=0.95\linewidth,keepaspectratio]{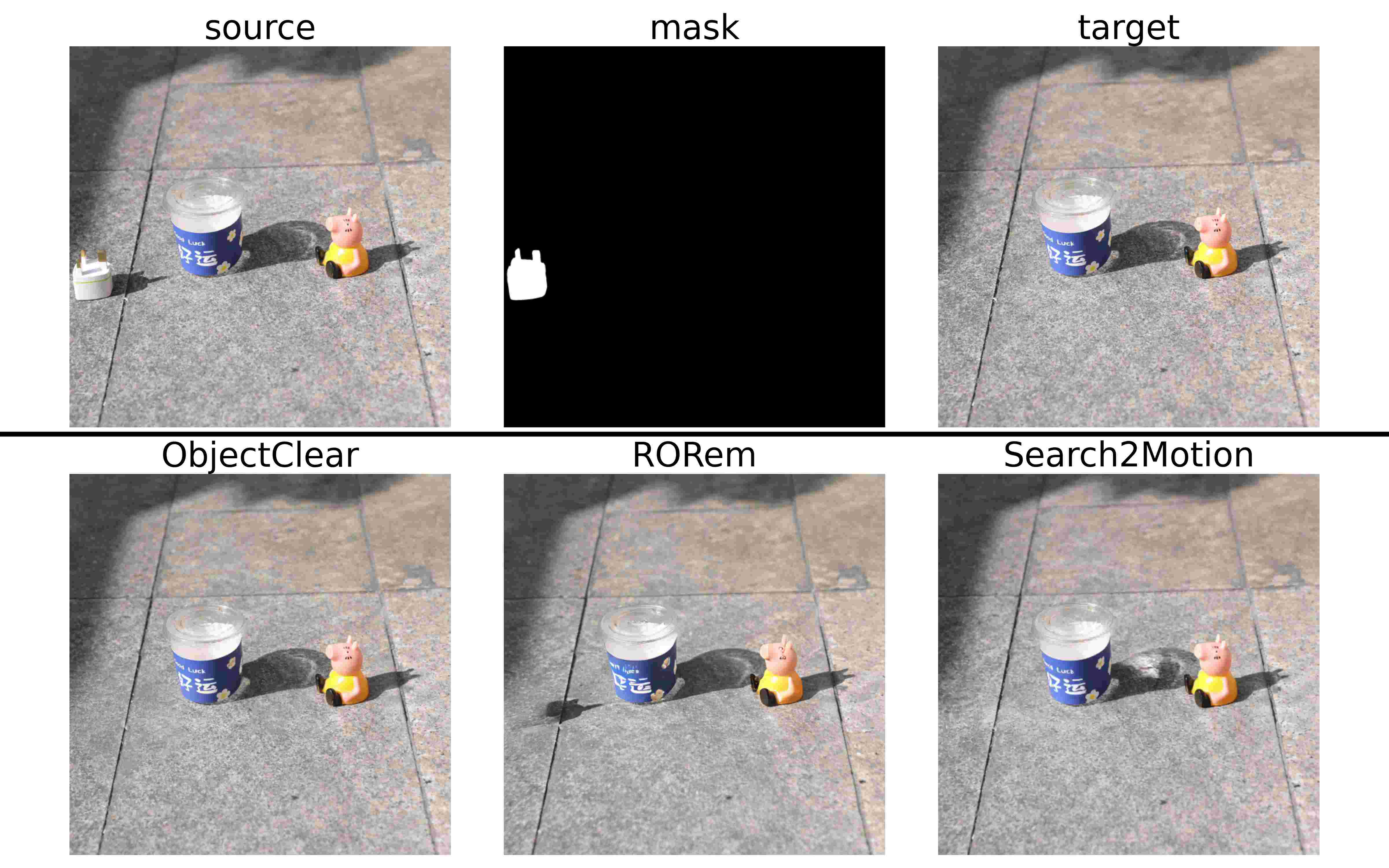}
    \vspace{-5pt}
    \captionof{figure}{Qualitative comparison of image object removal and background inpainting. The top row displays the source, mask, and target of sample $\text{real}\_001$ from the ObjectMover set A. The bottom row, from left to right, displays images generated by ObjectClear, RORem, and our inpainting methods. ObjectClear and our methods can remove shadows, whereas RORem does not.}
    \label{fig:background}
    \vspace{-5pt}
\end{figure}

We evaluate four candidate methods: two in-house training-free approaches built upon Generative Omnimatte~\cite{lee2025generative-omnimatte} and ROSE~\cite{miao2025roseremoveobjectseffects}, and two off-the-shelf methods, RORem~\cite{li2025RORem} and ObjectClear~\cite{zhao2025ObjectClear}. We assess them on the RORem dataset using standard inpainting metrics: PSNR, SSIM, FID, LPIPS, and fMSE (\cref{tab:background}), and complement this with qualitative comparisons on scenes with moving objects (\cref{fig:background}).

Qualitatively, our in-house methods and ObjectClear both produce smoother fill-to-boundary transitions and cleaner shadow removal compared to RORem, which exhibits perceptible artifacts particularly around moving objects. Quantitatively, our in-house methods achieve scores competitive with RORem, but underperform ObjectClear due to occasional pixel loss and sensitivity to text-prompt quality in the object removal step. ObjectClear demonstrates consistently stronger robustness across scene types.

Based on this analysis, we adopt ObjectClear as the background inpainting module in Search2Motion, prioritizing robustness and artifact-free output as the primary criteria for downstream video quality.





\begin{table}[t!]
    \centering
    \captionof{table}{Image object removal and background inpainting with $512\times512$ resolution. We evaluate these methods on the ObjectMover\cite{yu2025objectmovergenerativeobjectmovement} dataset. More specifically, the ObjMove-A set, which contains 200 (source, mask, target) samples.}
    \label{tab:background}
    \setlength{\tabcolsep}{5pt}
    \resizebox{\columnwidth}{!}{%
    \begin{tabular}{lcccccc}
    \hline
     & Training &  &  &  &  &  \\
    Method & Free & PSNR$\uparrow$ & SSIM$\uparrow$ & FID$\downarrow$ & LPIPS$\downarrow$ & fMSE$\downarrow$ \\
    \hline
    $\text{\ours{}}_{GenOmni}$ & \checkmark & 21.37 & \textbf{0.69} & 30.90 & 0.13 & 1160.80 \\
    $\text{\ours{}}_{ROSE}$ & \checkmark & \textbf{21.67} & 0.68 & \textbf{27.93} & \textbf{0.11} & \textbf{1148.89} \\
    \hline
    RORem\cite{li2025RORem} & - & 26.70 & 0.83 & 29.91 & 0.08 & 1181.25 \\    ObjectClear\cite{zhao2025ObjectClear} & - & \textbf{31.58} & \textbf{0.93} & \textbf{15.01} & \textbf{0.03} & \textbf{600.51} \\
    \hline
    \end{tabular}%
    }
    \vspace{5pt}
\end{table}


\subsection{\ours{} Benchmark Generation Pipeline}
\label{sec:supp_benchmark}
\begin{figure}[t!]
    \centering
    \includegraphics[width=\linewidth]{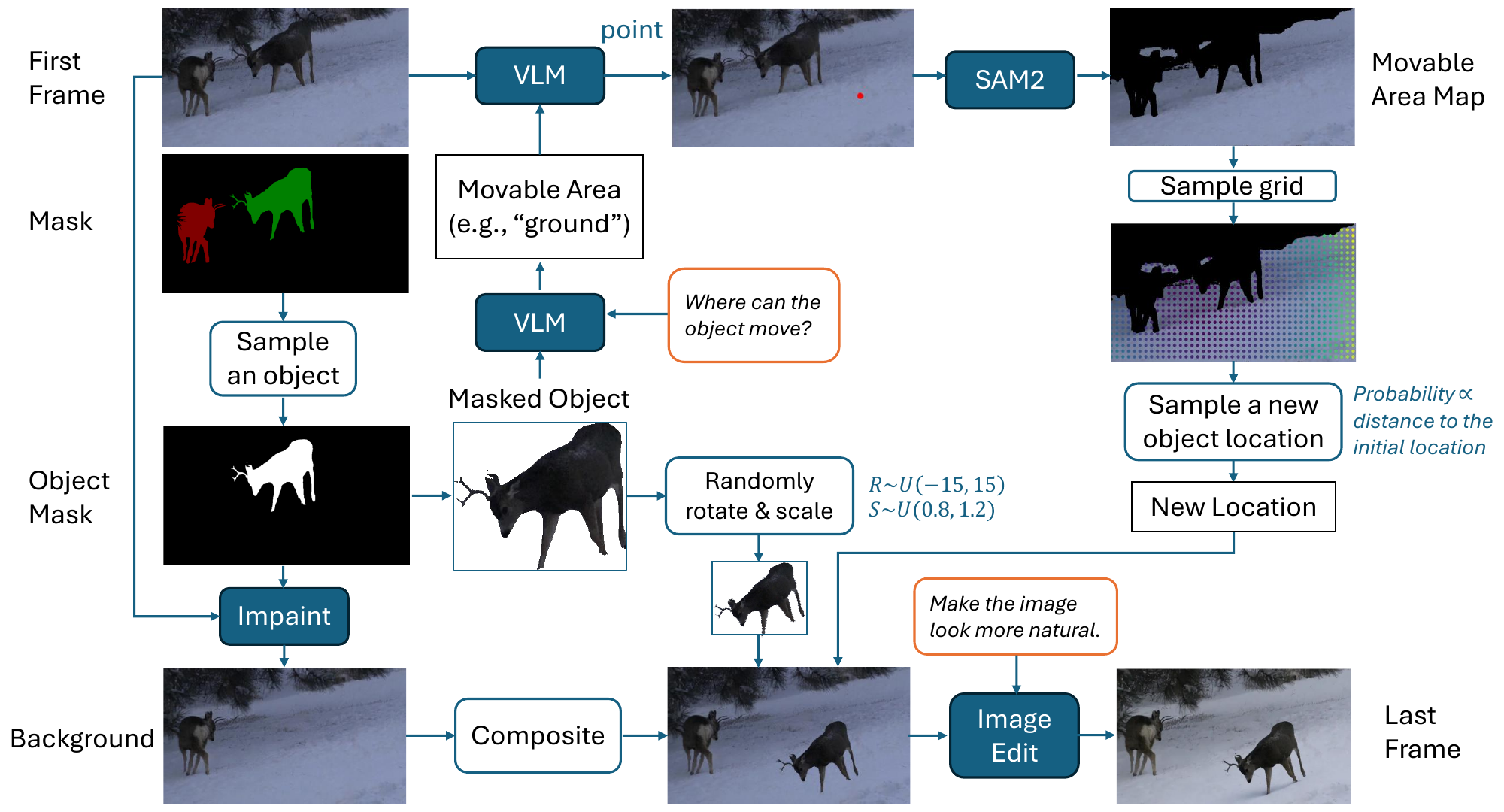}
    \captionof{figure}{Synthetic data generation for \davis{} through the semantic-guided object placement pipeline.}
    \label{fig:object-placement-pipe}
\end{figure}

\vspace{-5pt}
As described in Sec. 3.3, we synthesize the \davis{} benchmark using the semantic-guided object placement pipeline introduced in Sec. 3.1.
\cref{fig:object-placement-pipe} summarizes the process. We first extract an object with the ground truth mask, then determine the semantically valid placement regions in the scene using VLM and SAM2. A new object location is then sampled within such region using grid-based sampling, along with random rotation and scaling factors. Finally, we apply an image editing tool (e.g., Flux-Kontext) to harmonize the object with the background and enhance visual coherence.

\begin{figure}[t!]
    \centering
    \includegraphics[width=.95\linewidth]{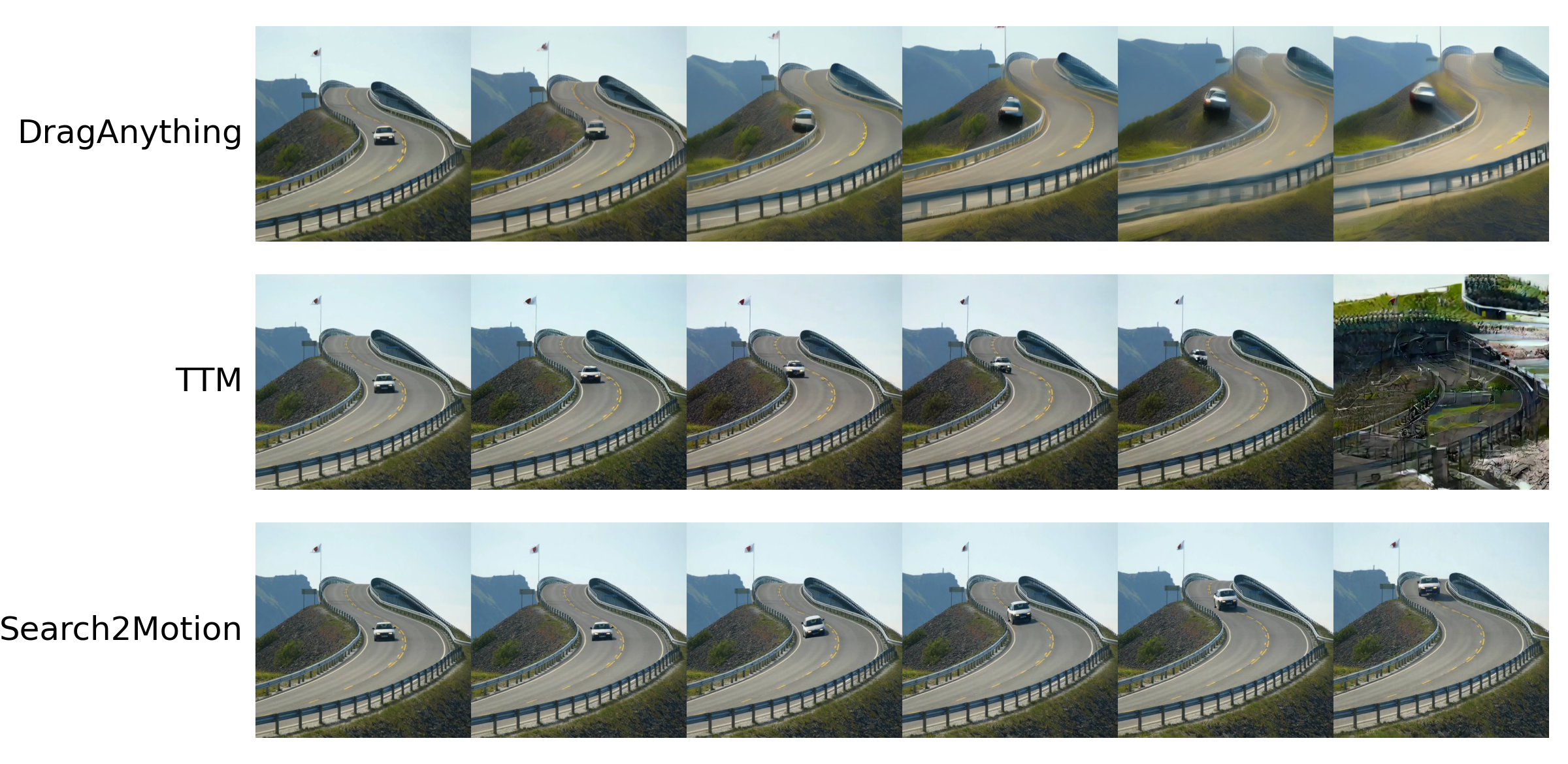}  \\\vspace{-5pt}
    \includegraphics[width=.95\linewidth]{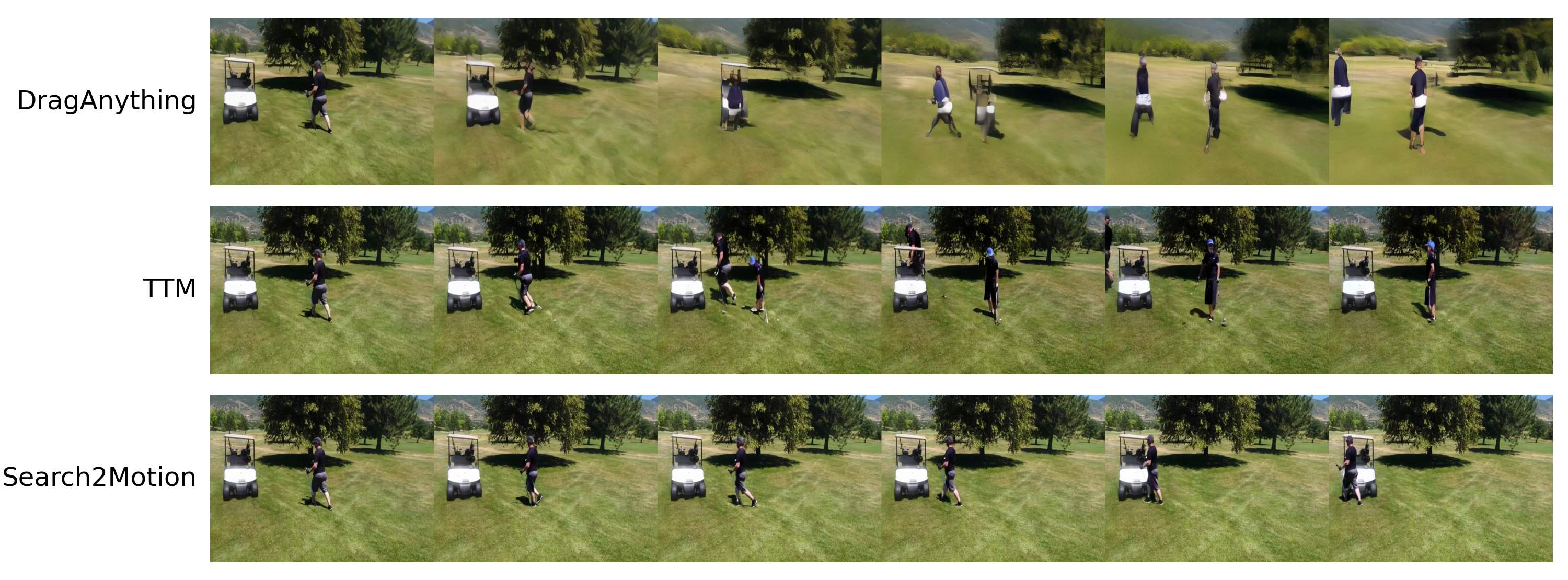}  \\\vspace{-5pt}\includegraphics[width=.95\linewidth]{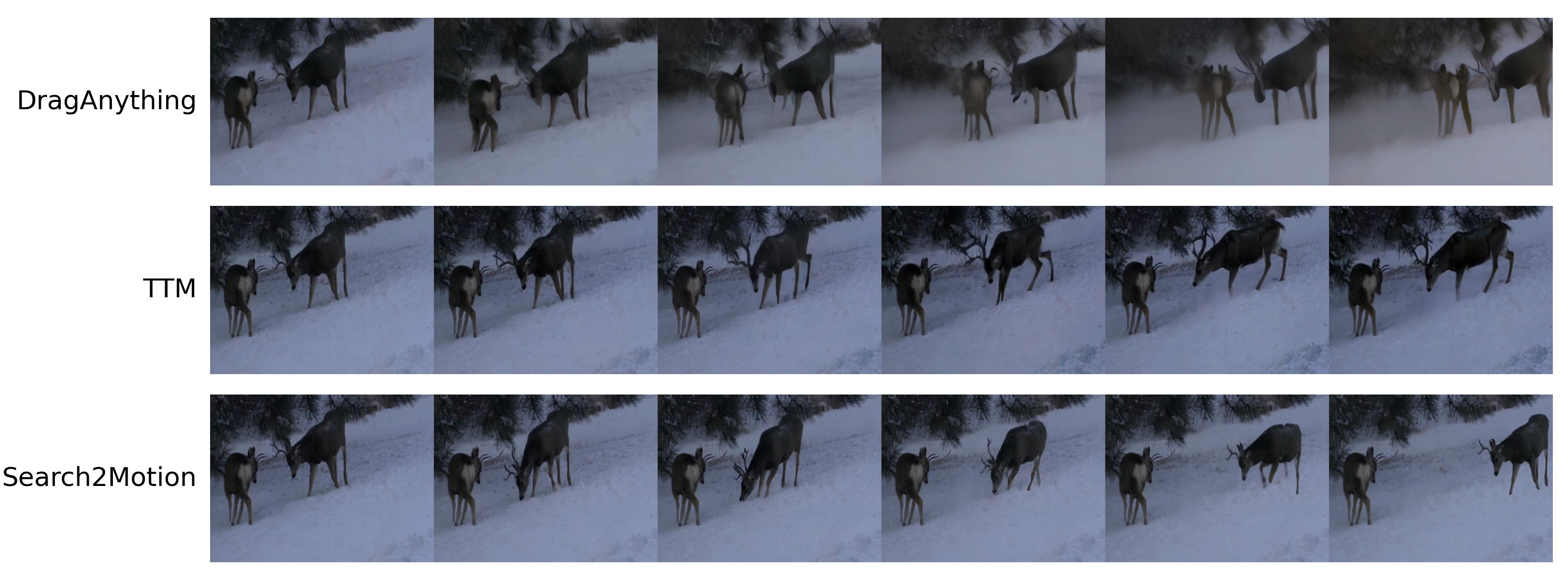}
    \vspace{-5pt}
    \captionof{figure}{Qualitative results of \ours{} video generation comparing against DragAnything and TTM on \objmove{} (top two) and \davis{} (bottom).}
    \label{fig:video_example_supp}
\end{figure}

\section{Experiments}
This section provides extended experimental results supplementing the main paper, including additional qualitative comparisons against trajectory-based control baselines (\cref{sec:supp_baselines}), visualizations of object end-state evaluation (\cref{sec:supp_vis_end_state_eval}), and further ablations on seed selection strategies for \search{} (\cref{sec:supp_aceseed}).
\subsection{Comparison to Trajectory-Based Control Baselines}
\label{sec:supp_baselines}

More qualitative examples are shown in \cref{fig:video_example_supp}, consistent with our discussions in Sec. 4.2. DragAnything often produces object distortion entangled with unintended camera motion. TTM generally maintains higher object fidelity but becomes unstable in later frames, and frequently fails to satisfy the last-frame condition. In contrast, \ours{} achieves stronger scene and object fidelity throughout the sequence and provides more precise control over the object's end state.

\subsection{Visualization of Object End State Evaluation}
\label{sec:supp_vis_end_state_eval}

\begin{figure}[t!]
    \centering
    \includegraphics[width=.82\linewidth]{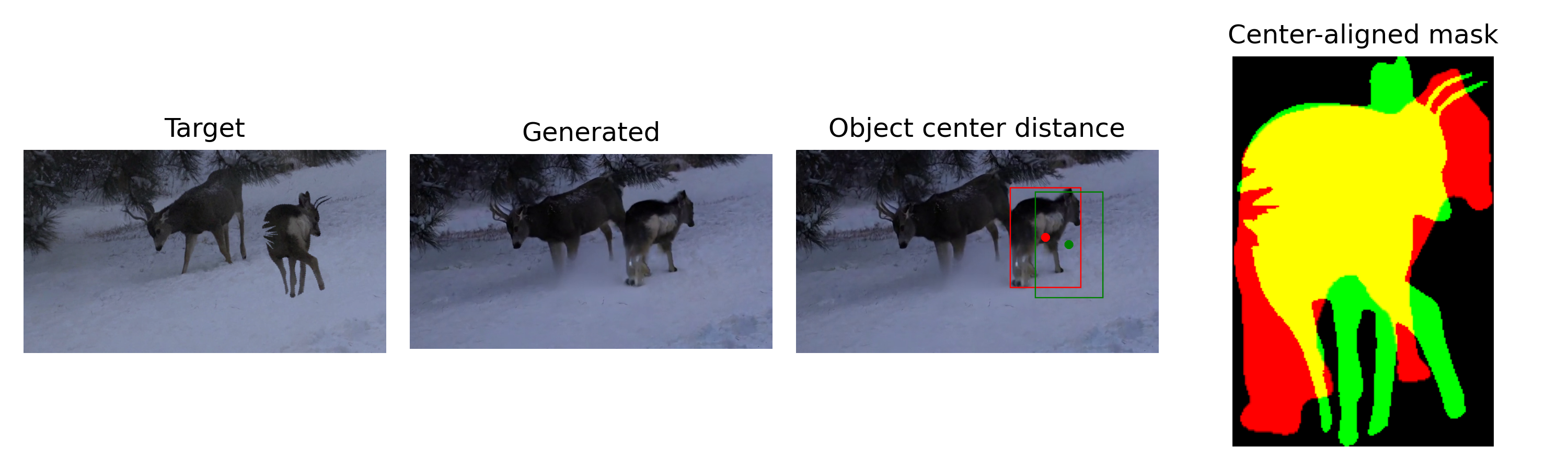}
    \includegraphics[width=.82\linewidth]{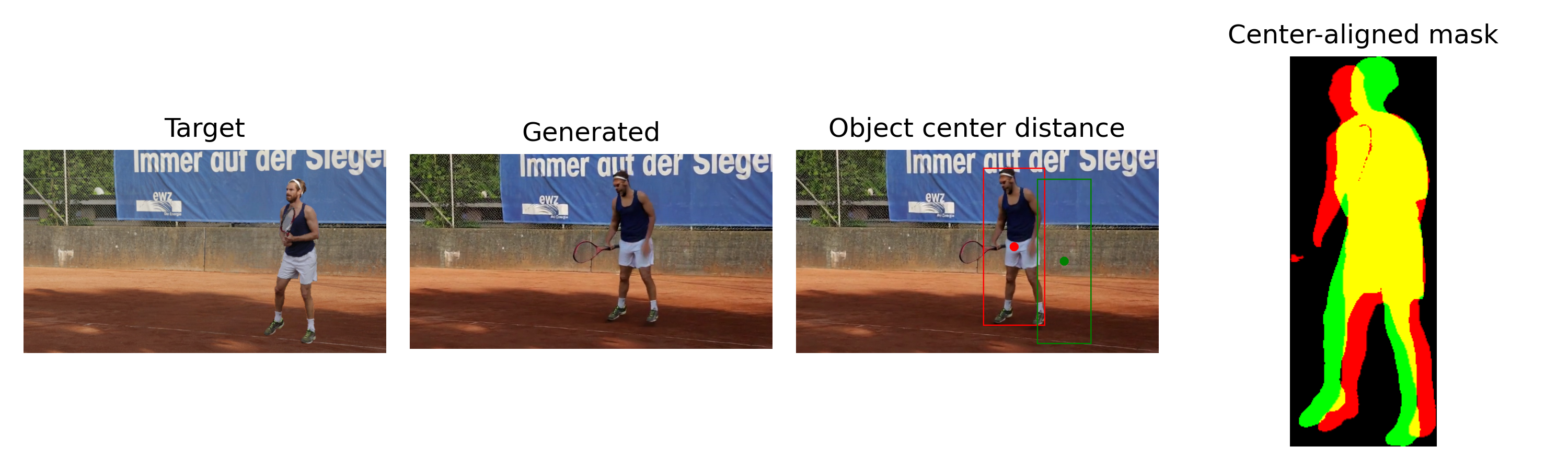}
    \includegraphics[width=.82\linewidth]{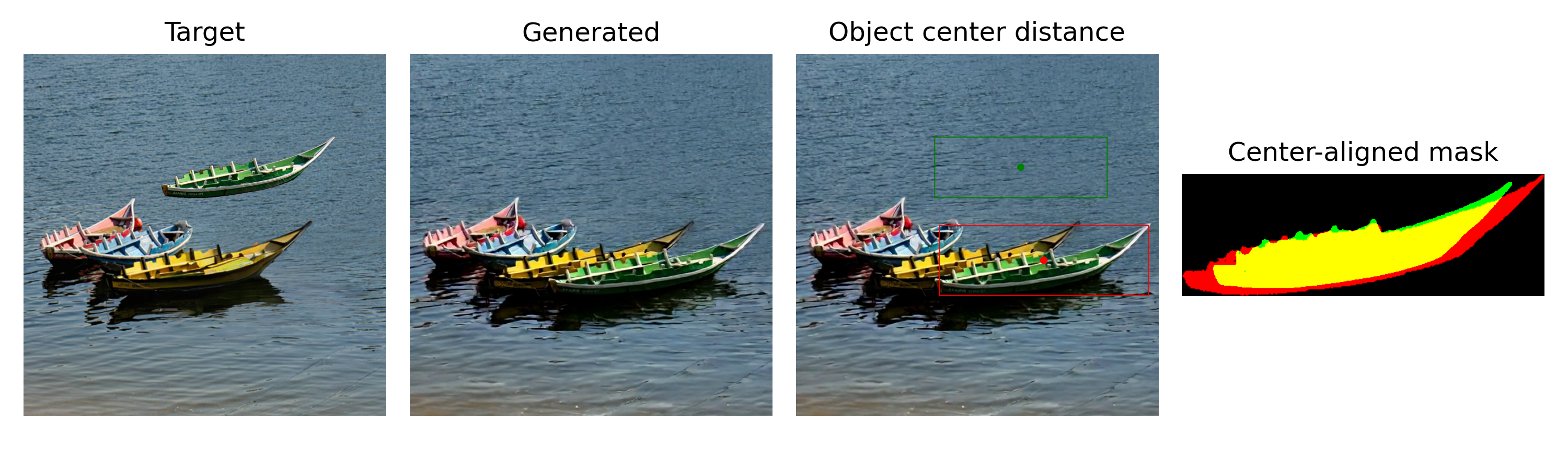}
    \vspace{-5pt}
    \caption{Object end state evaluation (green: ground truth; red: prediction).}
    \label{fig:obj_state_eval}
\end{figure}

While VBench and \metrics{} evaluate the whole-frame and object-centric consistency, respectively, we further assess object end state control with two metrics introduced in section 4.2: ObjMC and CA-IoU.
\cref{fig:obj_state_eval} shows visual examples comparing the generated and target ending frame. ObjMC measures how accurately the model places the object by computing the object-center distance between the predicted and target locations, independent of the object's pose. In contrast, CA-IoU evaluates pose fidelity by computing the IoU of the center-aligned masks, removing location differences. For example, in the last sample of \cref{fig:obj_state_eval}, the generated and target boat would yield an IoU of zero if location differences were not factored out.

\subsection{Ablation on Seed Selection Strategies}
\label{sec:supp_aceseed}

\subsubsection{Effectiveness of Seed Selection.}
We ablated seed selection strategies using Wan2.2-5B in Sec. 4.3. \cref{tab:search_vace} presents the ablation using another architecture, VACE-1.3B, which shows a similar trend.

\begin{table}[t!]
    \centering
    \captionof{table}{Seed selection ablation on VBench and \metrics{} metrics using VACE-1.3B. $\dagger$ denotes the average performance of non-selected seeds.}
    \vspace{-5pt}
    \setlength{\tabcolsep}{3pt}
    \resizebox{\linewidth}{!}{
    \begin{tabular}{ll|cccc|cccc}
    \toprule
    Dataset & Seed Selection & Subject$\uparrow$ & Background$\uparrow$ & Temporal$\uparrow$ & Motion$\uparrow$ & DINOv2$\uparrow$ & DINOv2$\uparrow$ & LPIPS$\downarrow$ & LPIPS$\downarrow$ \\
    & & Consistency & Consistency & Flickering & Smoothness & & first frame &  & first frame \\
    \midrule
    \objmove{} & none$\dagger$ & $94.07$ & $95.51$ & $97.53$ & $98.53$ & $96.26$ & $82.28$ & $.0592$ & $.2229$ \\   
    & \search{} & ${\bf 94.77}$ & ${\bf 95.98}$ & ${\bf 97.99}$ & ${\bf 98.75}$ & {\bf 96.92} & {\bf 84.46} & {\bf .0498} & {\bf .1965}\\
    \midrule
    \davis{} & none$\dagger$ & $94.39$ & $95.14$ & $97.91$ & $98.69$ & $93.93$ & $76.39$ & $.0904$ & $.2849$ \\
    & \search{} & ${\bf 95.69}$ & ${\bf 95.89}$ & ${\bf 98.53}$ & ${\bf 99.00}$ &
    {\bf 95.19} & {\bf 81.16} & {\bf .0809} & {\bf .2644}\\
    \bottomrule
    \end{tabular}
}
\vspace{5pt}
    \label{tab:search_vace}
\end{table}

\subsubsection{Best-Worst Separation.}
\begin{figure}[t!]
    \centering
    \includegraphics[width=\textwidth]{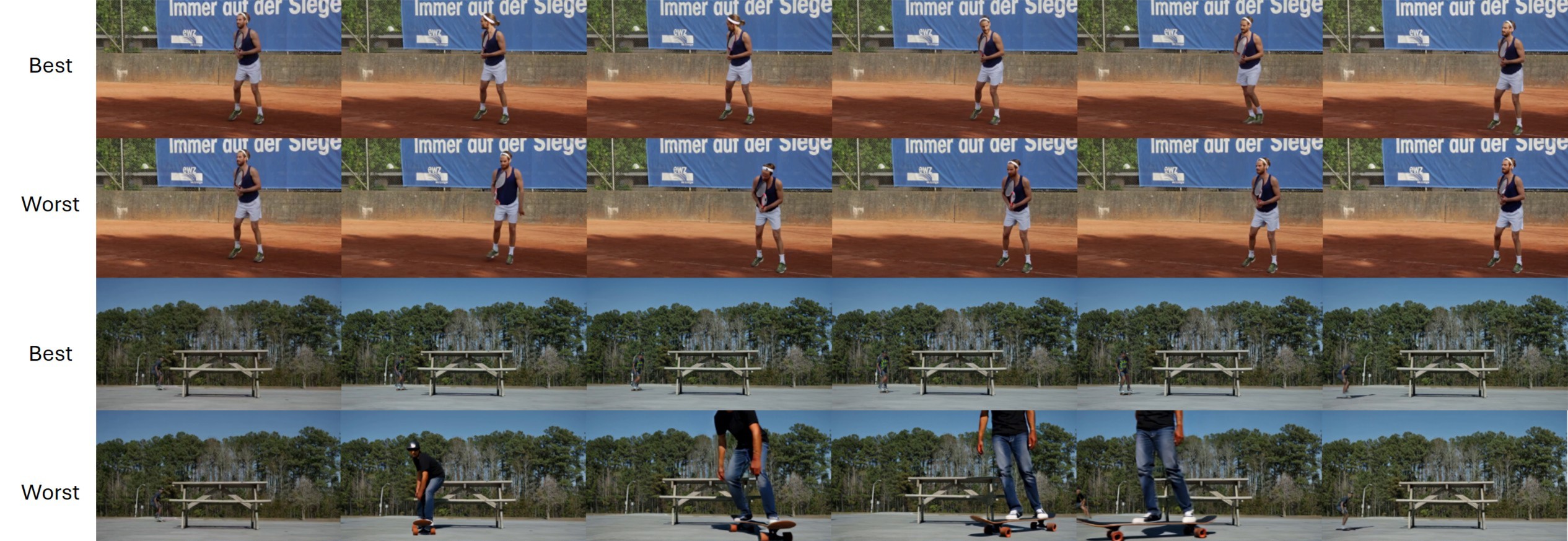}
    \caption{Additional qualitative Best-Worst comparison ranked by \search{} algorithm from the \davis{} dataset.}
    \label{fig:best_vs_worst_davis_supp}
\end{figure}

\begin{figure}[t!]
    \centering
    \includegraphics[width=0.9\textwidth]{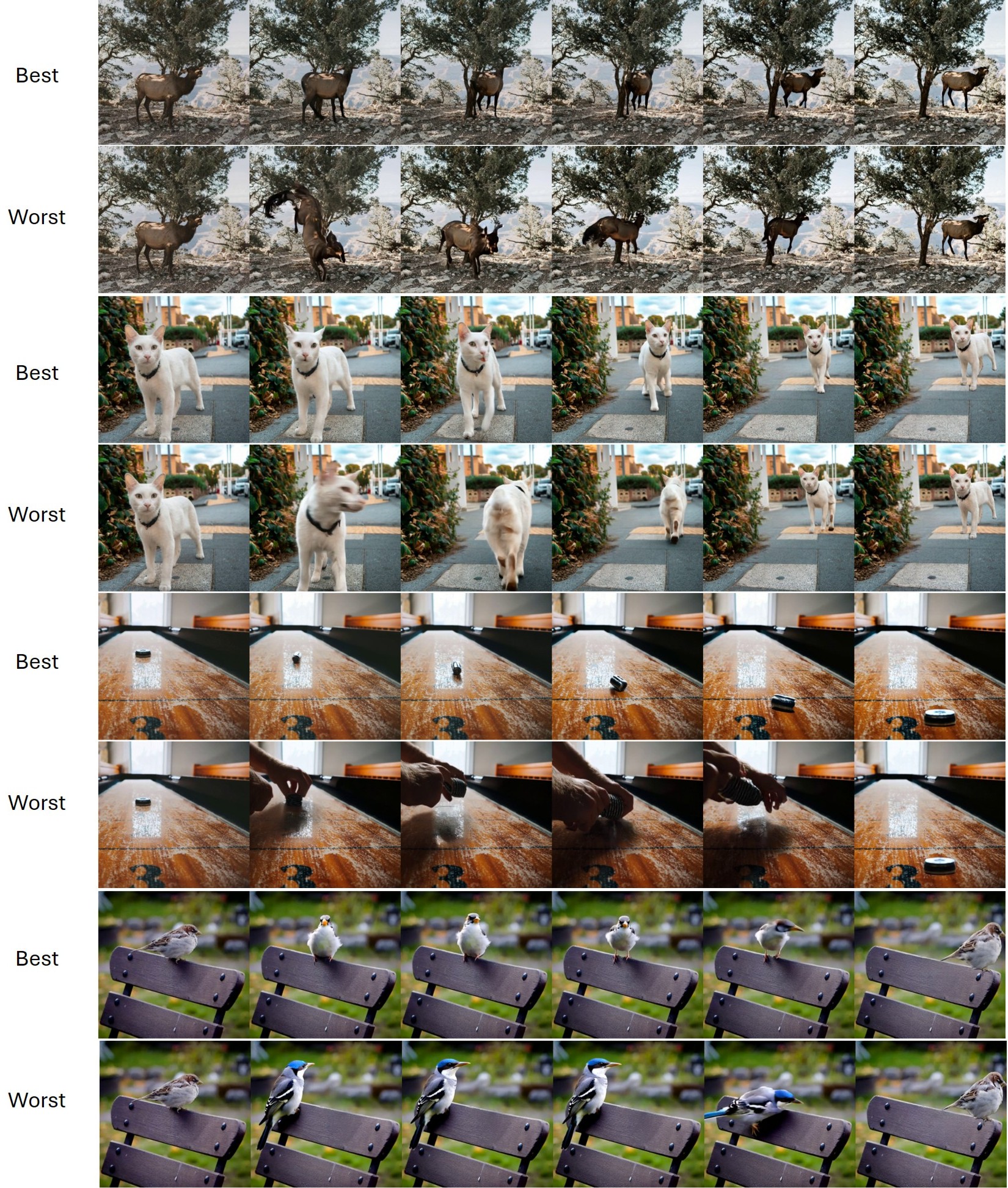}
    \caption{Additional qualitative Best-Worst comparison ranked by \search{} algorithm from the \objmove{} dataset.}
    \label{fig:best_vs_worst_objmove_supp}
\end{figure}

To further demonstrate the effectiveness of the proposed \search{} algorithm, we present additional examples in \cref{fig:best_vs_worst_davis_supp} and \cref{fig:best_vs_worst_objmove_supp},
showing the best and worst video outputs from a 10-seed search with examples from \davis{} and \objmove{} datasets, respectively. For each example, we use early-step attention consensus to score and rank all candidate generations. The resulting comparisons demonstrate that our algorithm consistently identifies the least plausible one, underscoring the discriminative power of early attention cues.

\subsubsection{Number of Seeds $N$ Ablation.}
\begin{figure}[t!]
    \centering
    \begin{subfigure}[h]{.49\linewidth}
        \centering
        \includegraphics[width=\linewidth]{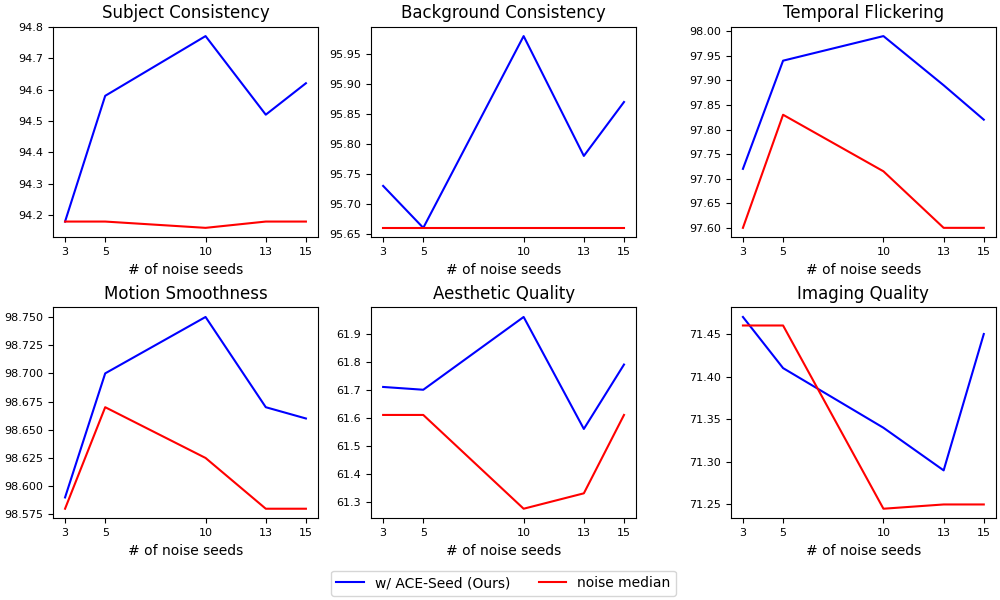}
        \caption{\objmove{}}
        \label{fig:objmove-b-nsample-vbench}
    \end{subfigure}
    \begin{subfigure}[h]{.49\linewidth}
        \centering
        \includegraphics[width=\linewidth]{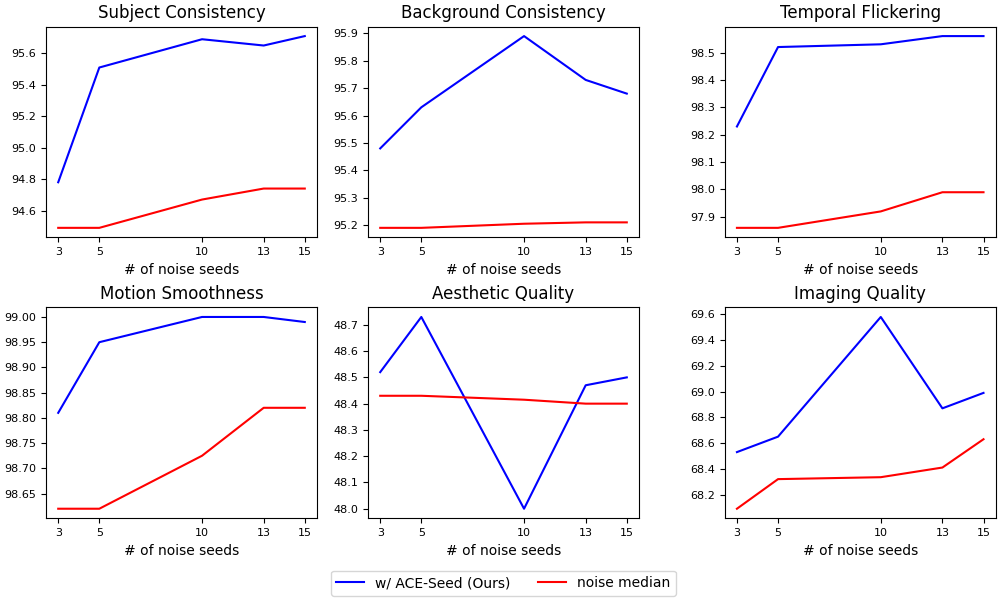}
        \caption{\davis{}}
        \label{fig:davis-nsample-vbench}
    \end{subfigure}
    \vspace{-5pt}
    \caption{VBench ablation for \search{} with different numbers of noise seeds. Higher is better.}\label{fig:nsample-vbench}
    \vspace{-5pt}
\end{figure}

\begin{figure}[t!]
    \centering
    \begin{subfigure}[h]{.49\linewidth}
        \centering
        \includegraphics[width=\linewidth]{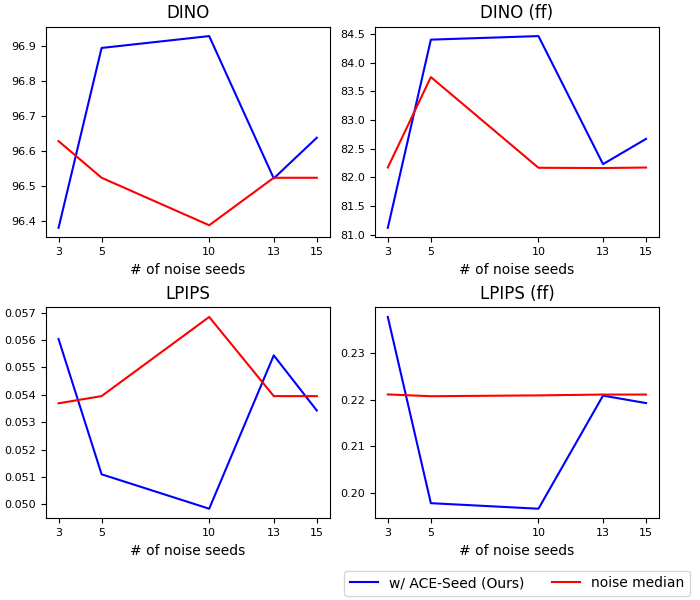}
        \caption{\objmove{}. Higher is better in the top row and lower is better in the bottom row.}
        \label{fig:objmove-b-nsample-ours}
    \end{subfigure}
    \begin{subfigure}[h]{.49\linewidth}
        \centering
        \includegraphics[width=\linewidth]{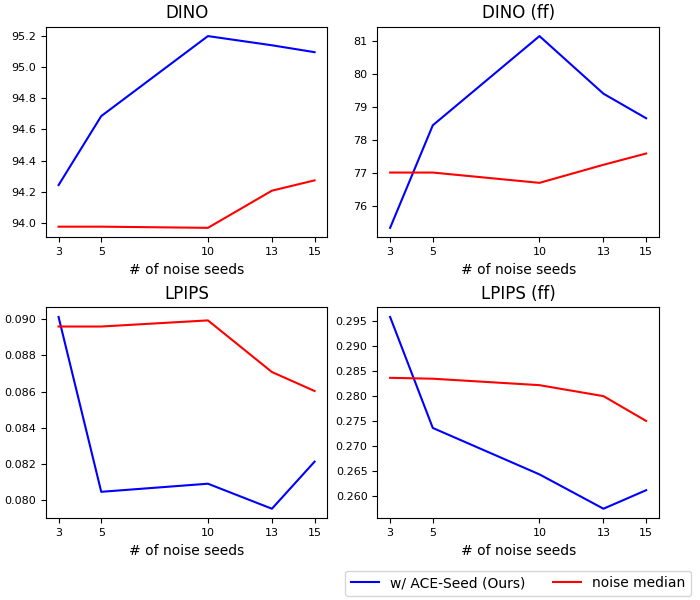}
        \caption{\davis{}. Higher is better in the top row and lower is better in the bottom row.}
        \label{fig:davis-nsample-ours}
    \end{subfigure}
    \caption{\metrics{} ablation for \search{} with different numbers of noise seeds.}\label{fig:nsample-ours}
    \vspace{-5pt}
\end{figure}

\cref{fig:nsample-vbench} and \cref{fig:nsample-ours} show that a higher number of seeds will result in better stability, but show diminishing returns beyond 13 to 15 seeds in both datasets. 


\begin{table}[t!]
    \centering
    \captionof{table}{VBench and \metrics{} ablation results on \davis{} for attention maps of different diffusion steps used in \search{}. All experiments share the same set of noise seeds. \colorbox{best}{Best}, \colorbox{second}{Second Best}.}
    \setlength{\tabcolsep}{3pt}
    \resizebox{\linewidth}{!}{
    \begin{tabular}{l|cccc|cccc}
    \toprule
    Step & Subject$\uparrow$ & Background$\uparrow$ & Temporal$\uparrow$ & Motion$\uparrow$ & DINOv2$\uparrow$ & DINOv2$\uparrow$ & LPIPS$\downarrow$ & LPIPS$\downarrow$\\
    & Consistency & Consistency & Flickering & Smoothness &  & First Frame &  & First Frame\\
    \midrule
    1 & 95.44 &	95.52 &	98.39 &	98.94 & 94.76 & 78.02 & 0.0834 & 0.2831\\
    4 & 95.35 &	95.47 &	98.43 &	98.94 & \cellcolor{second}95.06 & 79.11 & \cellcolor{best}0.0787 & \cellcolor{second}0.2705\\
    7 & \cellcolor{second}95.53 &	\cellcolor{second}95.82 &	\cellcolor{second}98.49 &	\cellcolor{second}98.98 & 94.93 & \cellcolor{second}80.48 & 0.0824 & 0.2723 \\
    10 & \cellcolor{best}95.69 &	\cellcolor{best}95.89 &	\cellcolor{best}98.53 &	\cellcolor{best}99.00 & \cellcolor{best}95.19 & \cellcolor{best}81.16 & \cellcolor{second}0.0809 & \cellcolor{best}0.2644\\
    \bottomrule
    \end{tabular}
}
\vspace{5pt}
    \label{tab:step_ablation}
\end{table}

\subsubsection{Attention Map Diffusion Step Ablation.}\label{sec:step_ablation}

To understand how spatial structure emerges during the diffusion process, we visualize attention overlays on the final generated video at sampled steps 1-13, 16, 30, and 50, as shown in \cref{fig:step_ablation}. Interestingly, coarse layout information begins to appear as early as step 1, even though the underlying latent is still dominated by noise. However, object boundaries and spatial relationships remain ambiguous during these earliest iterations. As the denoising process progresses, the attention fields sharpen rapidly, and by step 10 (out of 50 steps), the attention maps begin to provide the clearest and most stable indication of the final object layout, well before the visual content becomes recognizable. These observations form the foundation for our search algorithm, motivating the use of early-step attention consensus to identify the most reliable seeds. Additionally, we show that using attention maps at step 10 yields the best search results, as quantified in \cref{tab:step_ablation}.

\subsubsection{Attention Layer Selection.}
\begin{figure}[t!]
    \centering
    \includegraphics[width=\linewidth]{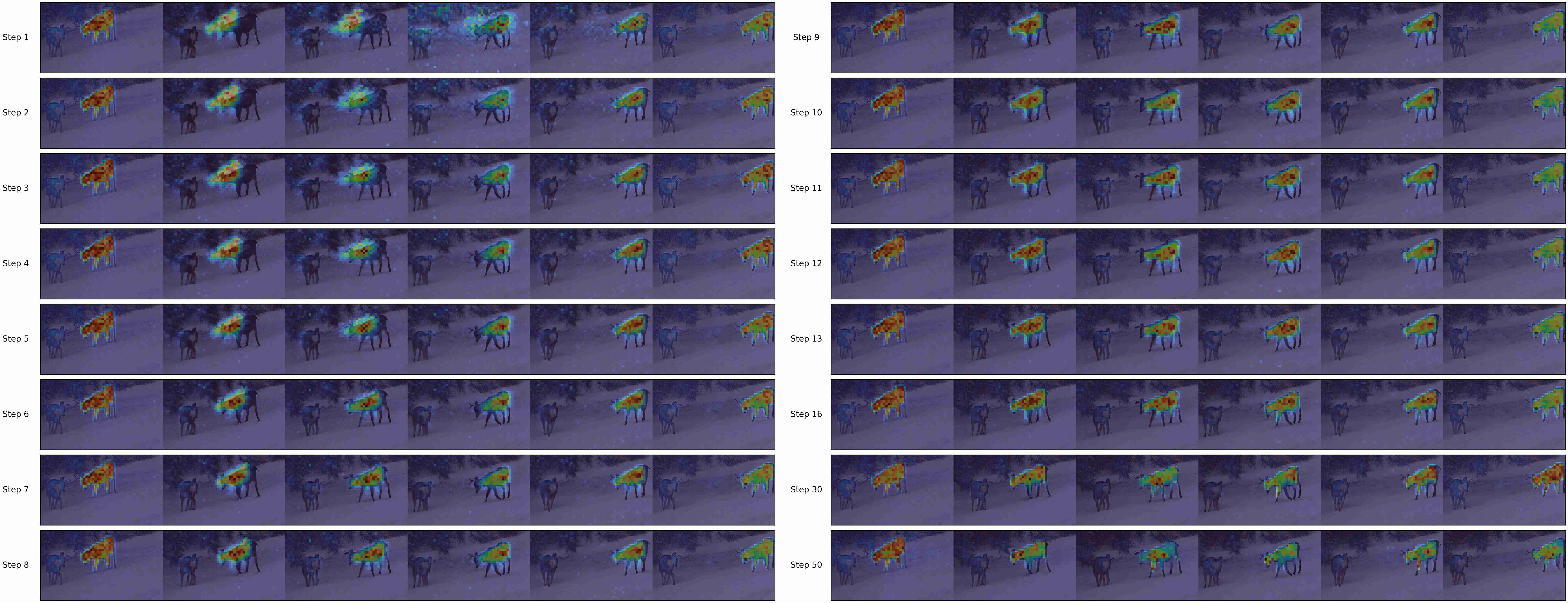}
    \vspace{-5pt}
    \captionof{figure}{Attention overlays on the final generated video across sampled denoising steps (ranges from 1 to 50). The attention maps begin to resemble the final object layout very early in the diffusion process, with the strongest structural alignment emerging around step 10.}
    \label{fig:step_ablation}
\end{figure}

\begin{figure}[t!]
    \centering
    \begin{minipage}{0.9\textwidth} %
        \centering
        \subfloat[Step 1 self-attention map qualitative ablation for all layers]{%
            \includegraphics[width=0.9\textwidth]{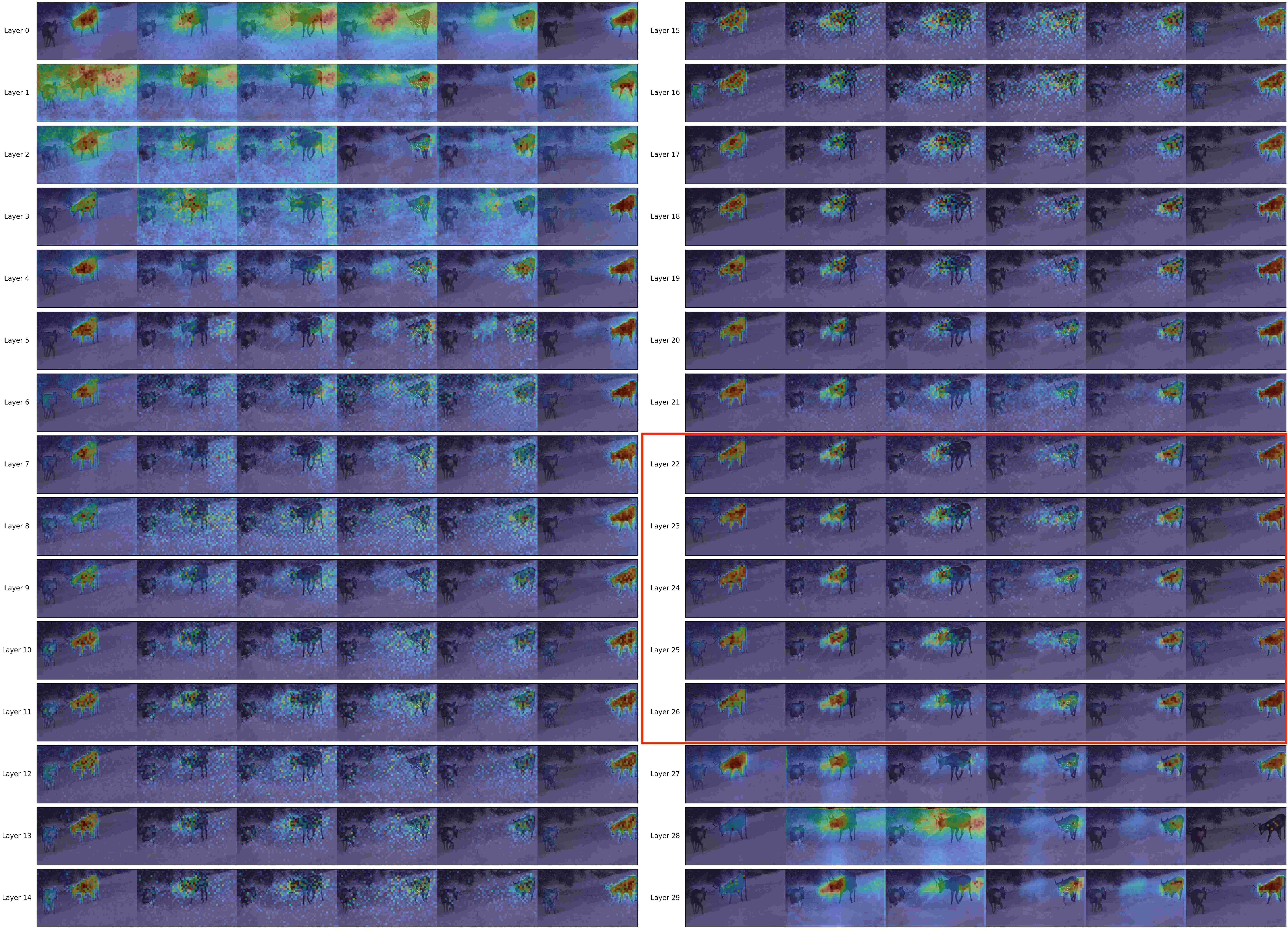}
            }\\[1ex]
        \subfloat[Step 10 self-attention map qualitative ablation for all layers]{
            \includegraphics[width=0.9\textwidth]{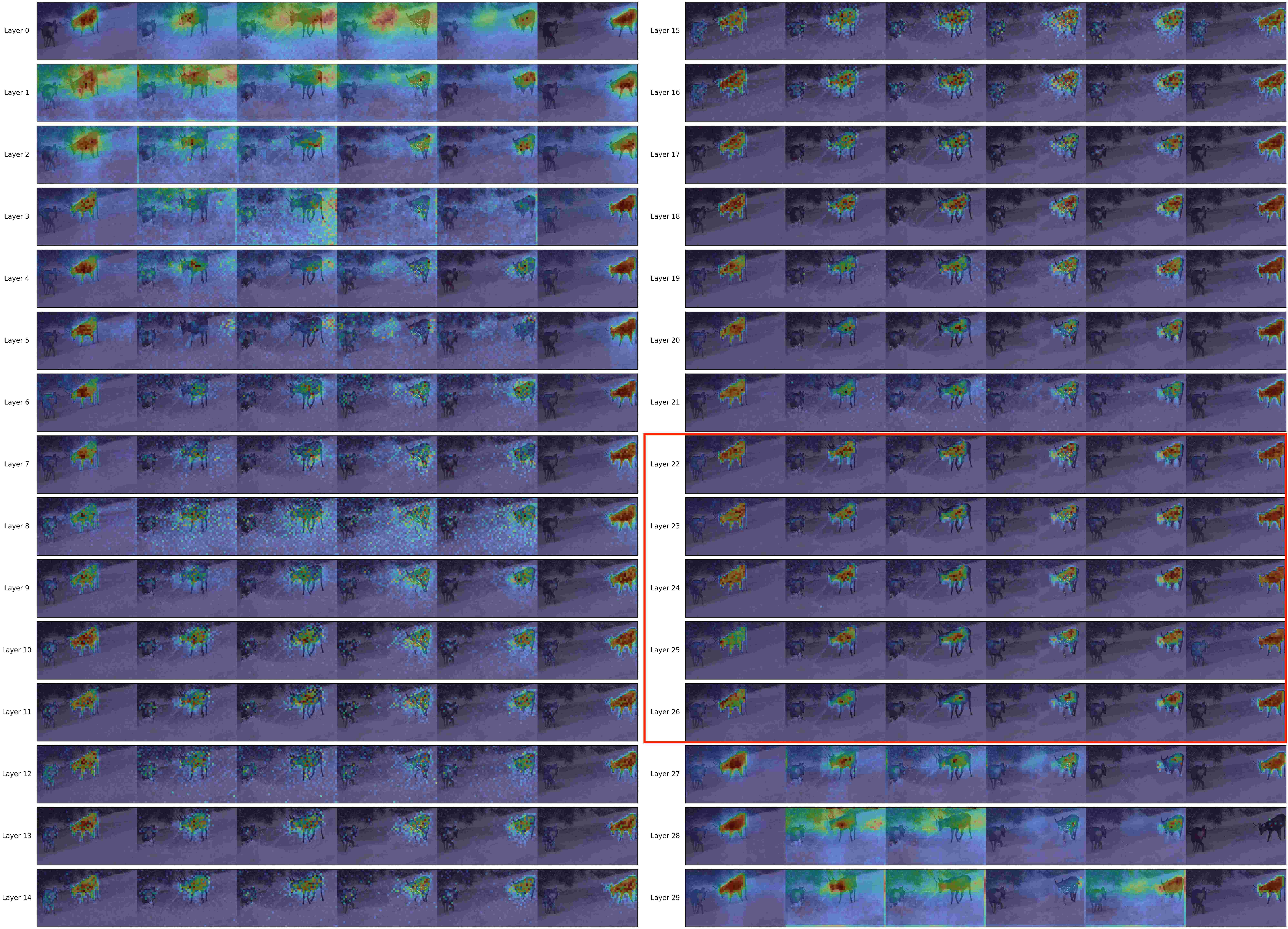}
            }
    \end{minipage}
    \caption{Self-attention overlays on the final video across attention blocks 0 to 29, shown for denoising step 1 (top) and step 10 (bottom). For both steps, blocks 22-26 produce the cleanest and most spatially coherent attention maps relative to other layers, given the first-frame foreground-mask tokens. Compared with step 1, the attention maps at step 10 exhibit noticeably clearer object contours that align better with the final video layout.}
    \label{fig:layer_ablation}
\end{figure}

To examine how different attention blocks contribute to spatial localization during early denoising, we visualize attention map overlays across blocks 0-29 for both denoising steps 1 and 10 in \cref{fig:layer_ablation}. Across both steps, we observe that blocks 22 to 26 consistently produce the least noisy and most spatially organized attention patterns, exhibiting clearer correspondence with the ground-truth foreground mask derived from the first frame. In contrast, lower- and mid-level layers show substantially noisier, more diffuse activations. Additionally, the comparison between step 1 and step 10 reveals that step 10 yields notably sharper and more complete object contours than step 1, reflecting closer alignment with the final video layout even though the denoising process is still in its early stages (consistent with the findings in \cref{fig:step_ablation}). These findings highlight a concentrated band of layers (22-26) that encode the most reliable spatial cues and further support our use of early-step, layer-aware attention consensus for seed selection. 

\begin{figure}[t!]
    \centering
        \includegraphics[width=.8\linewidth]{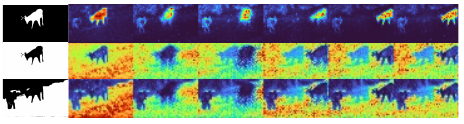}
         \vspace{-10pt}
        \captionof{figure}{Early-stage self-attention visualization under different token-masking strategies from S2M-DAVIS dataset. For each sample, from top to bottom: attention computed with the \textbf{foreground (fg) object mask}, \textbf{inverse foreground mask (bg)}, \textbf{placement mask (pl)}. It shows that bg can provide the most rich information (comparing to fg) with a relatively more consistent area coverage among different samples (comparing to pl).}
    \label{fig:mask_ablation_supp}
\end{figure}

\subsubsection{Token Selection (fg vs bg vs placement).}
In the main paper, we presented early-stage self-attention visualizations under different token-masking strategies for the \objmove{} dataset. Here, we also present a visualization from S2M-DAVIS (\cref{fig:mask_ablation_supp}).



\end{document}